\documentclass[preprint,12pt]{elsarticle}

\usepackage{amssymb}
\usepackage{graphicx}
\usepackage{subfigure}
\usepackage{multirow}
\usepackage{amsmath}
\usepackage{geometry}
\usepackage{tabularx}
\usepackage{amsmath}          
\usepackage{amssymb}
\usepackage{bm}
\usepackage{cite}
\usepackage[title]{appendix}
\usepackage{xcolor}
\usepackage{booktabs}
\usepackage{comment}

\usepackage{nomencl} 
\makenomenclature

\journal{Elsevier}
\begin{document}

\begin{frontmatter}


\title{Statistical Parameterized Physics-Based Machine Learning Digital Twin Models for Laser Powder Bed Fusion Process}

\author[inst1]{Yangfan Li}
\author[inst2]{	Satyajit Mojumder}
\author[inst3]{Ye Lu}
\author[inst1]{Abdullah Al Amin}
\author[inst2]{Jiachen Guo}
\author[inst1]{Xiaoyu Xie}
\author[inst1] {Wei Chen}
\author[inst1]{Gregory J. Wagner}
\author[inst1] {Jian Cao}
\author[inst1]{Wing Kam Liu\corref{cor1}}
\ead{w-liu@northwestern.edu}
\cortext[cor1]{Co-corresponding author}
\affiliation[inst1]{organization={Department of Mechanical Engineering},
            addressline={Northwestern University}, 
            city={Evanston},
            postcode={60208}, 
            state={IL},
            country={USA}}
\affiliation[inst2]{organization={Theoretical and Applied Mechanics Program},
            addressline={Northwestern University}, 
            city={Evanston},
            postcode={60208}, 
            state={IL},
            country={USA}}
\affiliation[inst3]{organization={Department of Mechanical Engineering},
            addressline={University of Maryland Baltimore County}, 
            city={Baltimore},
            postcode={21250}, 
            state={MD},
            country={USA}}

\begin{abstract}

A digital twin (DT) is a virtual representation of physical process, products and/or systems that requires a high-fidelity computational model for continuous update through the integration of sensor data and user input. In the context of laser powder bed fusion (LPBF) additive manufacturing, a digital twin of the manufacturing process can offer predictions for the produced parts, diagnostics for manufacturing defects, as well as control capabilities. This paper introduces a parameterized physics-based digital twin (PPB-DT) for the statistical predictions of LPBF metal additive manufacturing process. We accomplish this by creating a high-fidelity computational model that accurately represents the melt pool phenomena and subsequently calibrating and validating it through controlled experiments. In PPB-DT, a mechanistic reduced-order method-driven stochastic calibration process is introduced, which enables the statistical predictions of the melt pool geometries and the identification of defects such as lack-of-fusion porosity and surface roughness, specifically for diagnostic applications. Leveraging data derived from this physics-based model and experiments, we have trained a machine learning-based digital twin (PPB-ML-DT) model for predicting, monitoring, and controlling melt pool geometries.  These proposed digital twin models can be employed for predictions, control, optimization, and quality assurance within the LPBF process, ultimately expediting product development and certification in LPBF-based metal additive manufacturing.

\end{abstract}




\end{frontmatter}


\printnomenclature


\nomenclature{$W$}{melt pool width (m)}
\nomenclature{$D$}{melt pool depth (m)}
\nomenclature{$We$}{experiment melt pool width (m)}
\nomenclature{$De$}{experiment melt pool depth (m)}
\nomenclature{$Ws$}{simulation melt pool width (m)}
\nomenclature{$Ds$}{simulation melt pool depth (m)}
\nomenclature{$K$}{Gaussian kernel}
\nomenclature{$n$}{number of sample points}
\nomenclature{$i$}{variable index}
\nomenclature{$j$}{variable index}
\nomenclature{$h$}{bandwidth}
\nomenclature{$m$}{mode index}
\nomenclature{$F$}{mode function}
\nomenclature{$M$}{number of modes}
\nomenclature{$q_{source}$}{laser heat (W)}
\nomenclature{$P$}{laser power (W)}
\nomenclature{$V$}{scan speed ($\mathrm{m \cdot s^{-1}}$)}
\nomenclature{$\eta$}{absorptivity}
\nomenclature{$r_b$}{laser source radius (m)}
\nomenclature{$d$}{laser source depth (m)}
\nomenclature{$z_{top}$}{z-coordinate of the top surface (m)}
\nomenclature{$x_b$}{x coordinates of local reference system (m)}
\nomenclature{$y_b$}{y coordinates of local reference system (m)}
\nomenclature{$P1$}{calibration parameter}
\nomenclature{$P2$}{calibration parameter}
\nomenclature{$P3$}{calibration parameter}
\nomenclature{$\boldsymbol{\Sigma}$}{covariance matrix}
\nomenclature{$e$}{linear energy density ($\mathrm{J \cdot m^{-1}}$)}
\nomenclature{$NED$}{normalized energy density}
\nomenclature{$VED$}{volumetric energy density ($\mathrm{J \cdot m^{-3}}$)}
\nomenclature{$NED$}{normalized energy density}
\nomenclature{$H$}{hatch spacing (m)}
\nomenclature{$L$}{layer thickness (m)}
\nomenclature{$\rho$}{material density ($\mathrm{kg \cdot m^{-3}}$)}
\nomenclature{$C_p$}{specific heat ($\mathrm{J \cdot kg^{-1} \cdot K^{-1}}$)}
\nomenclature{$T_l$}{melting temperature (K)}
\nomenclature{$T_0$}{ambient temperature (K)}
\nomenclature{$T$}{temperature (K)}
\nomenclature{$Sa$}{surface roughness (m)}
\nomenclature{$u$}{velocity ($\mathrm{m \cdot s^{-1}}$)}
\nomenclature{$t$}{time (s)}
\nomenclature{$\mu$}{dynamic viscosity ($\mathrm{Pa \cdot s}$)}
\nomenclature{$p$}{pressure (Pa)}
\nomenclature{$\beta$}{thermal expansion coefficient ($\mathrm{K}^{-1}$)}
\nomenclature{$g$}{gravitational acceleration ($\mathrm{m \cdot s^{-2}}$)}
\nomenclature{$B$}{numerical parameter}
\nomenclature{$f_l$}{volume fraction of liquid}
\nomenclature{$k$}{thermal conductivity ($\mathrm{W \cdot m^{-1} \cdot K^{-1}}$)}
\nomenclature{$\varepsilon$}{emissivity}
\nomenclature{$\sigma_s$}{Stefan-Boltzmann constant ($\mathrm{W \cdot m^{-2} \cdot K^{-4}}$)}
\nomenclature{$\gamma$}{surface tension ($\mathrm{N \cdot m^{-1}}$)}
\nomenclature{$\alpha$}{consolidation factor}
\nomenclature{$RHF$}{residual heat factor}
\nomenclature{$PPB-DT$}{parameterized physics-based digital twin}
\nomenclature{$PPB-ML-DT$}{parameterized physics-based machine learning digital twin}

\section{Introduction}
\label{sec:intro}
Recent advancements in manufacturing technologies and the rapid growth of computational power, storage capacity, and data accessibility have brought the concept of the digital twin (DT) to the forefront in the manufacturing domain. This DT concept bridges the physical realm with the digital domain through a digital model that requires continuous updates from real-world experiments and the establishment of a robust database. The Laser Powder Bed Fusion (LPBF) process in metal additive manufacturing (AM) has achieved significant success and has found extensive applications in the aerospace, automotive, and biomedical industries  \citep{doi:10.1080/00207540903479786,Yan2012EvaluationsOC,guo2013additive}, drawing considerable attention in the AM research community. The performance of parts manufactured through LPBF, such as fatigue, relies on selecting appropriate processing conditions to control structural defects like surface roughness and porosity. These defects often arise from improper melt pool formation, which can be attributed to insufficient energy absorption, leading to incomplete melting, or the presence of trapped gas resulting from vaporization \citep{cunningham2017synchrotron,tang2017prediction, yadollahi2017additive, mojumder2023linking}. Understanding the effects of processing conditions on these structural defects is crucial for producing reliable AM parts. A digital twin model of LPBF process can provide predictions, diagnostic capabilities for defects, and serve as a tool for online monitoring and defects mitigation by controlling the melt pool phenomena. Such a model can also help in comprehending the relationship between the manufacturing process and the resulting part's structure, which is essential for achieving the desired performance in a wide range of applications. 

In the LPBF process, the solid powder undergoes a transformation into liquid state through the application of intense localized heat from a laser. This process inherently carries uncertainty process due to variations in the process parameters, such as laser power, scan speed, fluctuation in boundary temperatures,  etc.\citep{hu2017uncertainty,li2019improved, moges2019review}. The material parameters (e.g, powder conductivity, absorptivity, etc.) are also a major source of uncertainty.  As the laser moves away, the liquid cools and solidifies resulting in the formation of a bulk material with a distinct microstructure. The formation of a melt pool during the laser scan is a crucial aspect that governs the interactions between solid powder materials \citep{witherell2014toward} and influences structural defects such as surface roughness and porosity. The cross-sectional area of the melt pool, typically characterized by its width and depth, indicates the formation of porosity, as inadequate overlap between melt pools can lead to increased porosity \citep{qiu2019influence}. By modeling these uncertain processing conditions, stochastic predictions of the melt pool geometries can provide a more accurate estimation of surface roughness and porosity defects.

Previously, deterministic thermal models based on finite element and finite volume methods have been used to model the transient AM process. The predictive accuracy of these models depends on the calibration of the heat source model and the choice of the material properties. For instance, Ghosh et al. \citep{ghosh2018single} developed a finite-volume-based simulation model that effectively captures the melt pool geometries under various laser power and scan speed combinations and validated against experimental data. However, many of these models often overlook the fluid flow within the melt pool, thereby neglecting the impact of cooling through fluid convection, resulting in reduced accuracy in predicting melt pool geometries compared to thermal-fluid flow models that incorporate fluid dynamics, as highlighted by Yan et al. \citep{yan2020data}. Notably, Gan et al. introduced a well-tested transient three-dimensional thermal-fluid computational model capable of predicting both the thermal field throughout the entire part and the velocity field within the melt pool region \citep{Gan2021}. Their model was calibrated using highly controlled experiments conducted during the Additive Manufacturing (AM) Modeling Challenge Series in 2020 \citep{schwalbach2021afrl}, ensuring its accuracy and reliability. However, while this model has demonstrated accurate predictions of melt pool geometries, its limitation lies in the absence of stochastic information which restricts the model's ability to predict surface defects such as surface roughness \citep{zhao2018effect} and volumetric structural defects like porosity \citep{zhang2021microstructure}. Powder scale simulations has also been considered to simulate surface roughness and porosity for a smaller region in a deterministic manner \citep{STRANO2013589, 10.1115/MSEC2018-6501, ning2020analytical}. However, these models are computationally expensive and hinder the inclusion of part scale effects, thereby preventing direct comparisons with experimental measurements conducted at an engineering scale. Moreover, the deterministic nature of these models, lacking stochastic information, further restricts the accuracy of predictions.

For online monitoring and control of the melt pool, a fast computational model is essential. While physics-based models can provide accurate predictions of the melt pool geometries and defects, the time required for such predictions is often taking tens of seconds for control applications, especially where rapid predictions (in milliseconds) are needed between printing each layer. Machine learning (ML) model, leveraging computational algorithms to analyze and interpret data, can aid on this fast predictions of melt pool dimensions by learning from an offline database. Liao et al. trained a simulation-guided ML model for the control application in the Direct Energy Deposition (DED) process \citep{liao2022simulation}. Kozjek et al. trained a random forest ML model for the LPBF process based solely on experimental data \citep{kozjek2022data}. Researchers have also explored Convolutional Neural Networks (CNNs)  to effectively and rapidly monitor melt pool dimensions due to their capacity to autonomously and dynamically acquire spatial hierarchies of features \citep{yuan2018machine,li2020quality,yang2019investigation}. The LPBF AM process is inherently time-dependent and embodies a sequential nature. The state of melt pool at any given moment relies on both the current processing parameter inputs and the historical data leading up to that point. Therefore, an ML model that can capture both spatial and the transient behavior of the melt pool following a process history is critical for predicting the process. In this paper, we applied a deep autoregressive network (DARN) for controlling melt pool dimensions \citep{gregor2014deep}. DARNs excel when compared to traditional feed forward neural networks (FFNNs) because they can capture not only the current state of the melt pool, as represented by its width and depth, but also the short-term temporal dependencies within the data that reflect the transient nature of the melt pool. DARNs are specifically designed to handle such temporal dependencies effectively, making them well-suited for predicting and controlling melt pool parameters.

To address the necessity of modeling the stochastic nature of the LPBF process, this paper introduces a statistical parameterized physics-based digital twin (PPB-DT) model. It accomplishes this through a stochastic calibration of the heat source model parameters, enabling statistical predictions of melt pool geometries and defects such as Lack of Fusion (LOF) porosity and surface roughness. The stochastic calibration of the heat source model of the LPBF process uses Higher-Order Proper Generalized Decomposition (HOPGD) \citep{2018IJNME.114.1438L,lu2019datadriven}, a specially designed tensor decomposition method for the learning of non-intrusive data and construction of reduced-order surrogate models. Experimental data \citep{schwalbach2021afrl}, processed from melt pool measurements (see Section \ref{sec2}, is utilized to calibrate the stochastic heat source model parameters and validate it. This PPB-DT model is applied for diagnosing NIST overhang part, especially for LOF porosity and surface roughness. For online monitoring and control applications, a machine learning digital twin model is trained. This physics-based machine learning digital twin (PPB-ML-DT) offers predictive capabilities for controlling the processing parameters for subsequent steps of LPBF tracks. In section \ref{sec2}, we describe the components of the PPB-DT  with stochastic calibration process for statistical predictions of the melt pool. Section \ref{sec3} demonstrates the capabilities of the PPB-DT model in statistically predicting melt pool geometries and diagnosing defects such as LOF porosity and surface roughness. The control applications using PPB-ML-DT model have been demonstrated in Section \ref{sec4}. Section \ref{sec5} and \ref{sec6} provide the discussion of the presented results and outline some possible directions for future research. Finally, a conclusion is followed in Section \ref{sec7}.

\section{Digital twin model development of laser powder bed fusion process}
\label{sec2}
\begin{figure}[h!]
\centering
\includegraphics[width=1\textwidth]{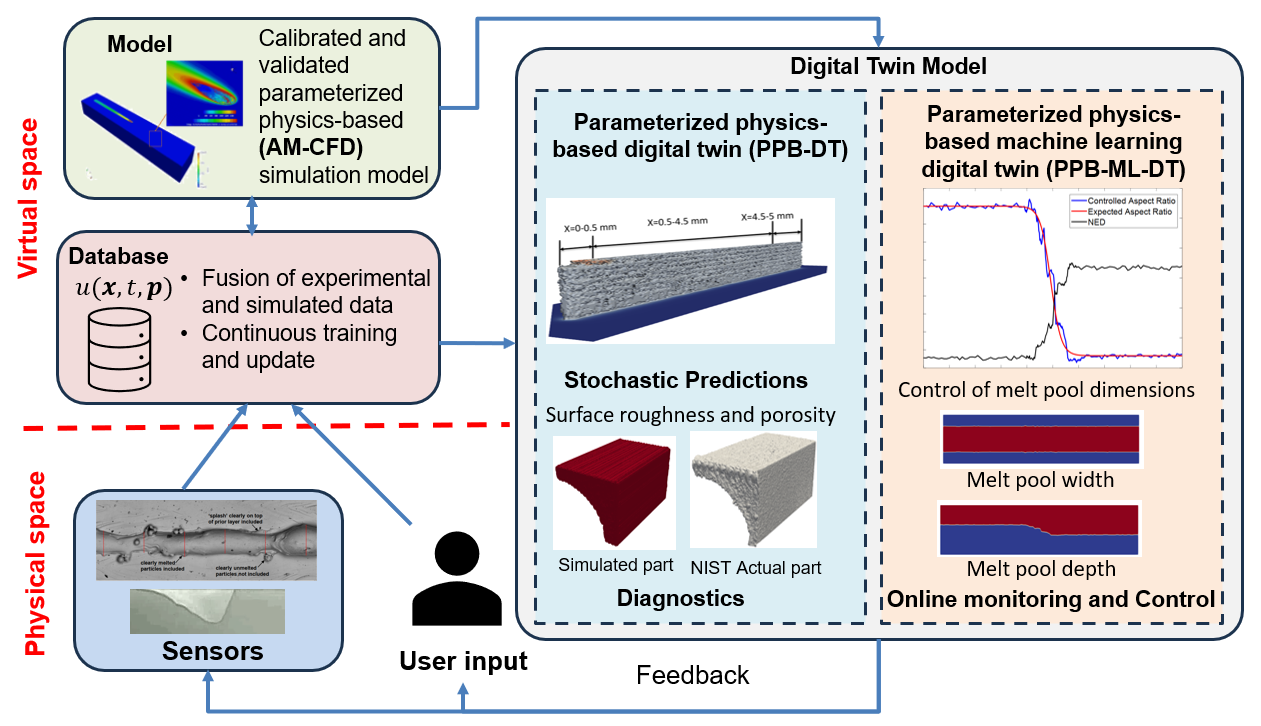}
\caption{Physics-based digital twin model development process and applications}
\label{fig1}
\end{figure}

A digital twin (DT) model operates on the intersection of the virtual and physical space by continuously updating the model with data from experiments and simulations. This process results in development of a database containing both the experimental and simulation data. The experimental data can be used for calibrating and validating the physics-based simulation model, and it can also be directly used in conjunction with the simulation data to construct the DT model. The calibrated and validated computational model serves as a parameterized physics-based digital twin (PPB-DT) model that is capable of predictive and diagnostic applications, as illustrated in Figure \ref{fig1}. However, certain applications, such as online monitoring and control, demand rapid prediction from the DT model. The PPB-DT model may not be suitable for such predictions, as it requires significant computational time, whereas the responses are needed in almost real-time. A potential solution for such applications lies in a  machine learning based model trained on the experimental and offline PPB-DT  model generated data. Utilizing the database developed through offline PPB-DT computation and experiment, a physics-based machine learning digital twin model (PPB-ML-DT) is trained to provide rapid prediction capabilities for the melt pool phenomena. 

In this section, we will describe the experimental data available for calibration and validation of our parameterized physics-based stochastic AM model. We will then outline the methodology for the stochastic calibration process, that enables the statistical predictions of the melt pool geometries, including the prediction of LOF porosity and surface roughness. Additionally, we will detail the process of developing the parameterized database and introduce the PPB-ML-DT model along with its training procedure. 

\subsection{Available experimental data for calibration and validation of physics-based stochastic AM model}
\label{sec2.1}
Laser powder bed fusion (LPBF) processes strongly trigger evaporation with complex gas flow which causes non-uniformity in the printed structure affecting the properties of the printed part. A validated computational model, obtained through a well-designed experiment, is essential for understanding the relationship between process, structure, and properties (PSP) and achieving desired performance in parts. In November 2019, the United States Air Force Research Laboratory: Materials and Manufacturing Directorate Structural Materials, Metals Branch (AFRL/RXCM) and America Make publicly announced the Additive Manufacturing Modeling Challenge Series. This initiative provided a series of highly controlled additive manufacturing experiments for validation and quantification of computational models \citep{Cox2021AFRLAM}.

\begin{figure}[h!]
	\centering
	\subfigure[Top-down description and melt pool width measurements]{
		\begin{minipage}{8cm} 
                       \includegraphics[width=\textwidth]{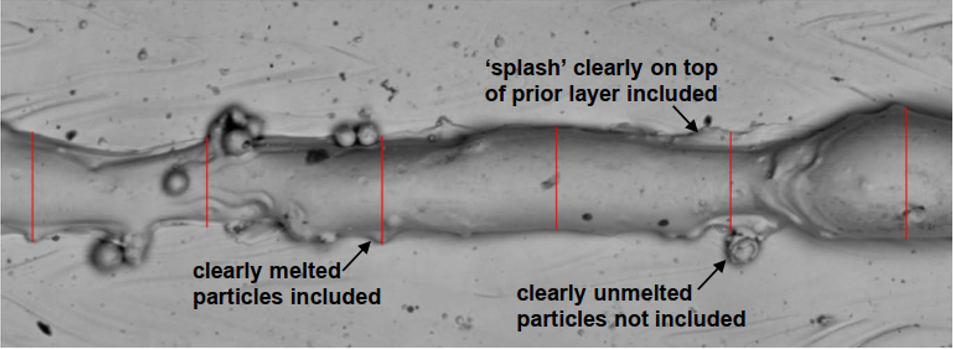}
		\end{minipage}
	}
    ~
	~	
	\subfigure[Top-down description and melt pool depth measurements]{
		\begin{minipage}{6cm}
			\includegraphics[width=\textwidth]{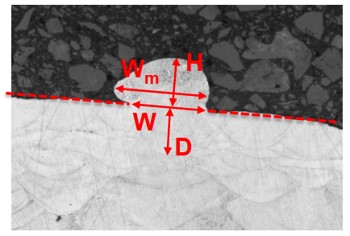} 
		\end{minipage}
  \label{fig2}
		}
\caption{AFRL experiment measurements \citep{Cox2021AFRLAM}: a) Top-down and b) cross-section melt pool description. In the top-down description, the red lines (Figure 2 a)) are samples of melt pool width measurements. In cross-section descriptions, $W$ is the width of melt pool, $W_m$ is the largest value of all widths, D and H are the depth and height of the deepest position of melt pool.}	
\end{figure}

In the AFRL experiment, different cases including single-layer single-track, single-layer multi-track, and multi-layer single-track (thin-wall) builds of IN625 powder are produced with an EOS M280 commercial LPBF system. Melt pool dimensions were measured using a electron back-scatter diffraction for top-down track description (Figure \ref{fig2}a) and optical microscopy on etched cross sections (Figure \ref{fig2}b). Detailed descriptions of the experimental setup and procedures can be found in the reference \citep{schwalbach2021afrl}. To accurately calibrate a stochastic AM model, single-track experiments are utilized to collect statistical measurements. Multi-track and multi-layer cases are then used to validate the melt pool geometry using the single-track calibrated model. Further, surface roughness and lack-of-fusion porosity are measured for the multi-track case and validated against experiments.

To determine the impact of heat source parameters on the melt pool size, it was crucial to analyze how these parameters affected the width ($W$) and depth ($D$) of the single-track melt pool. The measured value of the width and depth are shown in Table \ref{tab:table1} and \ref{tab:table2}, where $\mu$ represents the mean value and $\sigma$ denotes the standard deviation. Table 1 demonstrates the different measurements of melt pool width taken at various locations. The fourth column (20 locations) displays the results of the AFRL measurement, while the fifth to seventh columns (30 to 50 locations) show the measurements conducted in this study. Similarly, Table \ref{tab:table2} compares the melt pool depth between the AFRL measurements and this study. The last column, labeled "Depth", is the sum of the cross-section depth and height. 

\begin{table}[htbp]
  \centering
  \caption{Width $(\mu m)$ measurement for 11 single-track cases (A11-A11). AFRL conducted measurements at the 20 locations, while the additional measurements for this study, ranging from 30 to 50 locations, were based on the experimental images provided by AFRL.}
  \resizebox{\textwidth}{35mm}{
    \begin{tabular}{|c|c|c|c|c|c| p{3cm}|c|}
    \hline
    \multicolumn{1}{|p{3.89em}|}{Case } & \multicolumn{1}{p{4.61em}}{Laser } & \multicolumn{1}{|p{4.22em}|}{Scan} &       &       &       &  \\
    \multicolumn{1}{|p{3.89em}|}{number} & \multicolumn{1}{p{4.61em}}{Power } & \multicolumn{1}{|p{4.22em}|}{ Speed} & \multicolumn{1}{p{6em}|}{20 locations } & \multicolumn{1}{p{6em}|}{30 locations  } & \multicolumn{1}{p{6em}|}{40 locations} & \multicolumn{1}{p{6em}|}{\; 50 locations } \\
          & \multicolumn{1}{p{4.61em}}{($W$)} & \multicolumn{1}{|p{4.22em}|}{($mm/s$)} & $\mu$ \qquad $\sigma$ & $\mu$ \qquad $\sigma$ & $\mu$ \qquad $\sigma$ & \qquad $\mu$ \qquad $\sigma$\\
          \hline
    A1   & 300   & 1230  & 112.0±11.1 & 111.1±11.2 & 112.0±11.1 & \quad 111.9±10.9 \\
          \hline
    A2   & 300   & 1230  & 112.0±11.9 & 111.8±12.2 & 111.8±12.7 & \quad 111.3±12.1 \\
          \hline
    A3   & 290   & 953   & 127.6±7.0 & 124.7±9.1 & 125.5±10.5 & \quad 125.5±10.0 \\
          \hline
    A4   & 370   & 1230  & 122.9±8.4 & 119.4±10.0 & 117.7±10.2 & \quad 118.9±10.4 \\
          \hline
    A5   & 225   & 1230  & 96.0±13.9 & 100.1±14.2 & 99.7±13.1 & \quad 99.9±13.3 \\
          \hline
    A6   & 290   & 1588  & 97.9±14.0 & 99.7±11.3 & 100.7±13.6 & \quad 100.1±13.8 \\
          \hline
    A7   & 241   & 990   & 112.0±13.0 & 111.0±11.6 & 110.0±10.7 & \quad 109.4±10.5 \\
          \hline
    A8   & 349   & 1430  & 110.7±11.3 & 113.3±11.6 & 113.7±11.0 & \quad 113.4±11.3 \\
          \hline
    A9   & 300   & 1230  & 112.7±12.7 & 111.6±11.8 & 111.8±12.7 & \quad 112.2±11.8 \\
          \hline
    A10   & 349   & 1058  & 129.9±7.0 & 128.3±9.7 & 127.5±9.8 & \quad 127.3±9.4 \\
          \hline
    A11   & 241   & 1529  & 89.3±12.8 & 88.9±12.4 & 90.5±13.7 & \quad 90.8±13.4 \\
          \hline
    \end{tabular}%
  \label{tab:table1}}%
\end{table}%

\begin{table}[htbp]
  \centering
  \caption{Depth ($\mu$ m) measurement for 11 single-track cases (A1-A11). We compared our measurement with the AFRL measurements. Measurement data from this work is contrasted with the measurements from AFRL. In the final column, "Depth" represents the combined value of the cross-section depth and height.}
  \resizebox{\textwidth}{30mm}{
    \begin{tabular}{|c|c|c|c|c|c|c|p{1cm}|}
              \hline
    \multicolumn{1}{|p{4.055em}|}{Case } & \multicolumn{1}{p{4.055em}|}{Laser } & \multicolumn{1}{p{4.055em}|}{Scan} & \multicolumn{1}{p{6em}|}{Cross Section  } & \multicolumn{1}{p{6em}|}{Cross Section  } & \multicolumn{1}{p{6em}|}{Cross Section  } & \multicolumn{1}{p{6em}|}{Cross Section } &  \\
    \multicolumn{1}{|p{4.055em}|}{number} & \multicolumn{1}{p{4.055em}|}{Power } & \multicolumn{1}{p{4.055em}|}{ Speed} & \multicolumn{1}{p{7em}|}{Height (AFRL) } & \multicolumn{1}{p{8em}|}{Height (this work)  } & \multicolumn{1}{p{7em}|}{Depth (AFRL)  } & \multicolumn{1}{p{8em}|}{Depth (this work)  } & \multicolumn{1}{p{6em}|}{ Depth (sum) } \\
          & \multicolumn{1}{p{4.055em}|}{($W$)} & \multicolumn{1}{p{4.055em}|}{ ($mm/s$)} & $\mu$ \qquad $\sigma$ & $\mu$ \qquad $\sigma$ & $\mu$ \qquad $\sigma$ &  $\mu$ \qquad $\sigma$ & $\mu$   \hspace{0.6em}    $\sigma$ \\
                    \hline
    A1   & 300   & 1230  & 59.1±12.3 & 59.0±12.9 & 54.3±9.0 & 54.3±8.9 & 113.3±13.4 \\
              \hline
    A2   & 300   & 1230  & 65.7±21.8 & 65.7±21.7 & 52.3±9.0 & 52.5±8.6 & 118.2±19.9 \\
              \hline
    A3   & 290   & 953   & 68.1±9.2 & 68.1±9.1 & 72.0±7.4 & 72.0±7.4 & 140.0±12.8 \\
              \hline
    A4   & 370   & 1230  & 66.0±15.5 & 66.2±15.3 & 75.9±7.6 & 75.9±7.2 & 142.1±17.4 \\
              \hline
    A5   & 225   & 1230  & 60.3±14.9 & 60.3±14.9 & 25.0±6.1 & 25.0±6.1 & 85.3±13.6 \\
              \hline
    A6   & 290   & 1588  & 62.2±18.3 & 62.2±18.4 & 26.9±5.4 & 27.1±5.6 & 89.3±19.9 \\
              \hline
    A7   & 241   & 990   & 61.2±11.9 & 61.2±11.9 & 42.5±6.6 & 42.6±7.2 & 103.8±13.2 \\
              \hline
    A8   & 349   & 1430  & 60.1±15.9 & 60.1±16.1 & 58.5±4.6 & 58.5±4.6 & 118.5±18.2 \\
              \hline
    A9   & 300   & 1230  & 68.8±25.9 & 68.8±26.0 & 46.9±9.3 & 46.8±8.8 & 115.5±30.6 \\
              \hline
    A10   & 349   & 1058  & 63.5±17.8 & 63.3±17.6 & 84.0±8.9 & 83.8±8.6 & 147.1±19.4 \\
              \hline
    A11   & 241   & 1529  & 56.3±18.1 & 56.3±18.3 & 20.1±7.1 & 20.1±7.1 & 76.4±22.1 \\
              \hline
    \end{tabular}%
  \label{tab:table2}}%
\end{table}%

\subsection{Calibrated and validated parameterized physics-based digital twin (PPB-DT) model}

\subsubsection{Stochastic calibration of the heat source parameters of the physics-based AM model}
A physics-based model of the LPBF process can capture relevant melt pool phenomena, such as capillary and Marangoni flow, and keyhole formation. However, capturing all the melt pool physics in a model is not feasible and is computationally challenging, and many of these parameters are uncertain. Calibration provides an effective way to account for the unaccounted physics in the model and improve predictions.

Developing a good calibration model requires controlled experiments, and having a large experimental dataset can significantly enhance the model's accuracy in this regard. Predicting surface defects (e.g., surface roughness \citep{ZHAO201876}) and volumetric defects (e.g., porosity \citep{QU2022110454}) can be improved using a calibrated physics-based model. The accuracy of deterministic simulations in predicting defects, such as surface roughness and porosity, is limited, as it heavily relies on the quality of the model calibration.

\begin{figure}[h!]
\centering
\includegraphics[width=1\textwidth]{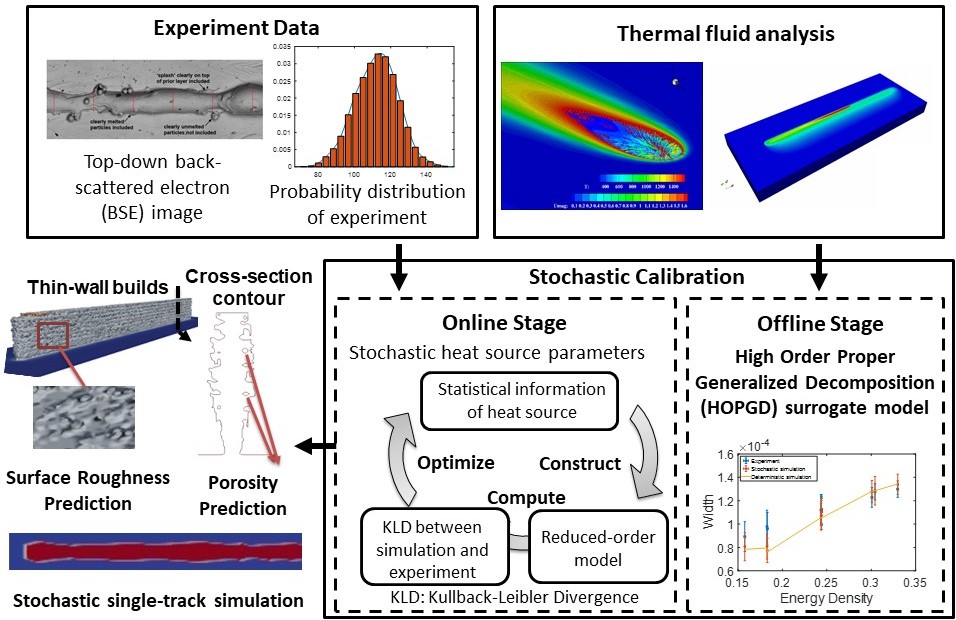}
\caption{Stochastic additive manufacturing simulation framework for LPBF process following a stochastic calibration of the heat source model parameters}
\label{fig:3}
\end{figure}

To address this issue, we propose a stochastic calibration framework (see Figure \ref{fig:3}) aimed at calibrating the heat source parameters of the physics-based AM model using experimental observations of melt pool geometries. Stochastic AM simulations also allow us to predict surface roughness and porosity in as-built parts, facilitating comparisons with experimental observations.

The stochastic physics-based AM modeling framework consists of the following components: i) analyzing experimental melt pool geometry (width and depth) to develop a probabilistic model from observations, ii) creating a thermal-fluid simulation incorporating a statistical heat source model, iii) stochastically calibrating the heat source model parameters, and iv) stochastically predicting LPBF process melt pool phenomena and part-scale defects.
The melt pool geometries (depth, and width) are obtained by analyzing the controlled AFRL experiment described above (see section \ref{sec2.1}). Top-down and cross-section of the melt pool images for various processing conditions are analyzed and probability distribution model of the melt pool width ($W$) and depth ($D$) is formulated from the experimental observations. The thermal-fluid analysis is based on our in-house code called "AM-CFD". The AM-CFD code has been rigorously tested and confirmed for its accuracy through the 2022 NIST AM Bench challenge, achieving three first-place awards \citep{yeung2018implementation, yang2019investigation}. Additionally, its prowess was demonstrated in modeling challenges led by the Air Force Research Laboratory (AFRL), where it secured another first-place award \citep{cox2021afrl, schwalbach2021afrl}. \ref{app1} provides details about our stochastic physics-based AM-CFD model \citep{gan2019benchmark,Gan2021}.  In our framework as presented in Figure \ref{fig:3}, we introduce stochastic parameters for the heat source model in AM-CFD. These parameters account for the uncertainties caused by variations in real experimental conditions and provide a stochastic prediction of the melt pool geometry. In stochastic calibration process, Kernel density estimation \citep{Davis2011} is used to develop a non-parametric distribution of the melt pool geometry. The stochastic AM-CFD predicted melt pool geometry is then statistically compared with the experimental melt pool geometry using Kullback–Leibler divergence (KLD) \citep{10.1214/aoms/1177729694}. To significantly reduce the computational cost for multi-parametric calibration, HOPGD is used to handle the AM-CFD heat source model parameters calibration problem. The calibrated stochastic AM-CFD can then simulate part-scale samples using a Markov chain Monte Carlo (MCMC) method \citep{Hastings1970MonteCS} by sampling the calibrated heat source parameters in different time series, with results better than deterministic models. Through this stochastic modeling framework AM-CFD can predict the surface roughness and lack-of-fusion (LOF) porosity of the as-built parts by simulating multilayer-multitrack parts.  

\begin{figure}[h!]
\centering
\includegraphics[width=1\textwidth]{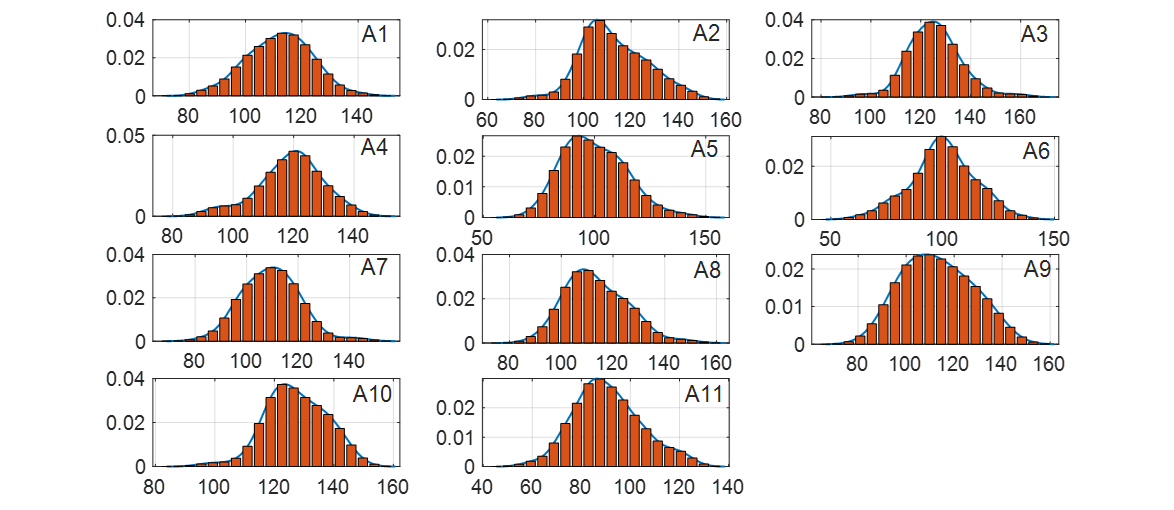}
\caption{Experimental measurement (orange bar) and its probability density function generated by KDE (blue line) of single experiment case A1 to A11. x-axis represents width, while y-axis is probability density.}
\label{Fig:KDE}
\end{figure}

To calibrate the heat source model, probability density functions (PDF) of experimental melt pool dimensions, width and depth, are calculated using Kernel Density Estimation (KDE)\citep{Davis2011}. KDE is a powerful method for estimating the PDF of a random variable. The distributions of the experimental and simulated melt pool width and depth are represented as follows:

 \begin{equation}
f_{We}(W)=\frac{1}{nh}\sum_{j=1}^{n}{K(\frac{W-W_{ej}}{h})}
\end{equation}

 \begin{equation}
f_{De}(D)=\frac{1}{nh}\sum_{j=1}^{n}{K(\frac{D-D_{ej}}{h})}
\end{equation}
 
 \begin{equation}
f_{Ws}(W)=\frac{1}{nh}\sum_{j=1}^{n}{K(\frac{W-W_{sj}}{h})}
\end{equation}

 \begin{equation}
f_{Ds}(D)=\frac{1}{nh}\sum_{j=1}^{n}{K(\frac{D-D_{sj}}{h})}
\end{equation}

\noindent where $f_{We}, f_{De},f_{Ws}, f_{Ds}$ represent distributions of experimental width ($We$), experimental depth ($De$), simulated width ($Ws$) and depth ($Ds$), respectively. $K$ is the Gaussian kernel, $j$ is the index of the sample point, and $n$ is the total number of sample points. $h$ represents the bandwidth. Detailed expression can be found in \ref{app2} along with a brief overview of KDE. The KDE results are shown in Figure \ref{Fig:KDE} for the 11 single-track experiment cases in PDF format. These experimental measurements will be used to calibrate the stochastic parameters of the heat source model in AM-CFD in the following section. 

Determining the appropriate heat source model parameters is crucial for achieving reliable simulation of the AM process. Multiple simulations are required to tune these calibration parameters in the model which leads to a significant computational burden. Traditional methods like the genetic algorithm \citep{whitley1994genetic}involve repeated calls to the computational model to evaluate the model parameters. To minimize computational expenses, we have integrated the HOPGD into our model. HOPGD is a non-intrusive surrogate model that utilizes a database. The database can be either from simulations or experiments, and the foundation behind proper generalized decomposition (PGD) is the separation of variables technique. This approach accelerates the calibration (identification) process for the heat source model parameters. For a n-dimensional function $f\left(\mu_1,\mu_2,\ \mu_{3\ \ldots.,\ }\mu_n\right)$ that contains the quantity of interest as a function of n parameters, HOPGD separation form is given by:

 \begin{equation}
f(\mu_1,\mu_2, \mu_3 \ldots.,\mu_n)\approx\sum_{m=1}^{M}{{F}_1^{(m)}(\mu_1){F}_2^{(m)}(\mu_2)}{F}_3^{(m)}(\mu_3)\ldots.,{F}_n^{(m)}(\mu_n)
\end{equation}

The function $f$ is given by the finite sum of products of the separated functions ${F}_{i}^{\left(m\right)}(i=1,..,n)$. ${F}_{i}^{\left(m\right)}$ identifies the variation of function ${f}$ in the parameter direction $\mu_i$, which is also called mode function. $n$ is the rank of approximation and $m$ defines the mode number of each component (not represent exponential terms). $M$ represents the total number of modes. The number of $n$ is priori unknown and can be obtained with a precomputed physics-based simulation database \citep{2018IJNME.114.1438L,lu2019datadriven,lu2018multi,lu2018space}. HOPGD seeks the projection of data for computing the mode functions that can reproduce the original function. This enables HOPGD to serve as a surrogate model for efficient prediction. The surrogate model reduces the computational cost since it only requires 1D interpolation to find output at a given point using the computed mode functions ${F}_{i}^{\left(m\right)}$. The method is suitable for high-dimensional problems due to parse sampling strategy, which is more challenging for other approaches. HOPGD has been applied successfully to accelerate the calibration of welding \citep{lu2019datadriven}, AM process \citep{gan2021benchmark}, and microstructural \citep{saha2021microscale} models under deterministic settings. Code examples for the HOPGD can be found on the GitHub project (https://yelu-git.github.io/hopgd/).

In this work, we extended the HOPGD method to a stochastic calibration setting. We consider a cylindrical heat source model \citep{amin2023physics,gan2019benchmark, gan2021benchmark, mojumder2023linking} to model the heat input by the laser which is given by: \begin{equation}
q_{source}=\begin{cases}                                   

                   \frac{2P \eta}{\pi r_b^2d}\exp{\left(\frac{-2\left(x_b^2+y_b^2\right)}{r_b^2}\right)}\ \ \ z_{top}-z\le\ d;\\            

                   0\ \ \ \ \  \ \ \ \ \ \ \ \ \ \ {\ \ \ \ \ \ \ \ \ \ \ \ \ \ z}_{top}-z \textgreater\ d\\             

                   \end{cases}
\end{equation}

 \noindent where $P$ denotes the laser power, $\eta$ is the absorptivity, $r_b$ is the laser beam radius, $d$ is the depth of the heat source, and $z_{top}$ is the z-coordinate of the top surface of the computational domain. $x_b$ and $y_b$ are the coordinates in the local reference system attached to the moving heat source. Note that the parameters, $\eta$, $r_b$ and $d$, are all unknown and uncertain heat source parameters, which are highly correlated to the vapor depression phenomenon in the LPBF process. During calibration, the minimum value of absorptivity is limited to 0.28\citep{osti_1569671}. According to the literature \citep{osti_1569671,2018JLasA..30c2410F}, it has been observed that increasing laser power or decreasing scan speed results in the formation of a vapor-induced depression and increases absorptivity through deepening the keyhole region by multiple reflections of the laser beam between the liquid and gas interface.  Therefore, we assume the three parameters, $\eta$, $r_b$ and $d$ are related to the laser power to scan speed ratio $P/V$, as follows:
 
 \begin{equation}
 d=P1\frac{P}{V}
\end{equation}

 \begin{equation}
\eta=\max(P2\frac{P}{V},0.28)
\end{equation}

 \begin{equation}
 r_b=P3\frac{P}{V}
\end{equation}
$P_1, P_2, P_3$ are considered as random parameters and calibrated with the information extracted from melt pool dimension data of AFRL experiments. We assumed that the variation of melt pool reflects the characteristic length of the surface roughness and lack of fusion porosity in the LPBF process. In this case, the stochastic heat source parameters are assumed to satisfy a tri-variate normal distribution:

 \begin{equation}
{{p}}\left(P1,P2,P3\right)=\frac{1}{{(2\pi)}^\frac{3}{2}\left|\boldsymbol{\Sigma}\right|^\frac{1}{2}}e^{-\frac{1}{2}\left[\left(\boldsymbol{P}-\boldsymbol{\mu}\right)^T\boldsymbol{\Sigma}^{-1}\left(\boldsymbol{P}-\boldsymbol{\mu}\right)\right]}
\end{equation}

\noindent where ${\boldsymbol{P}}=[P1,P2,P3]^ { T }$ is the vector of heat parameters, $\boldsymbol{\mu}=[\mu 1,\mu 2,\mu 3]^ { T }$ is the mean vector, and ${\boldsymbol{\Sigma}=\left[\begin {matrix}C11&C12&C13\\C21&C22&C23\\C31&C32&C33\\ \end{matrix}\right]}$ is the covariance matrix. Due to the symmetry of $\Sigma$, unknown coefficients are ${\mu_1,\mu_2,\mu_3,C}_{11},C_{22},C_{33},C_{12},C_{23},C_{13}$. These are the final uncertain hyper-parameters that need to be determined.


In our stochastic AM model, we seek to identify the relationship between heat source model and important parametric melt pool dimensions (width and depth). Hence, the HOPGD model reads

 \begin{equation}
{{Y}_{s}}={F}\left(e,P1,\ P2,\ P3\right)=\sum_{m=1}^{k}{{F}_1^{\left(m\right)}\left(e\right){F}_2^{\left(m\right)}\left(P1\right){F}_3^{\left(m\right)}\left(P2\right)}{F}_4^{\left(m\right)}\left(P3\right) \label{eq13}
\end{equation}
\noindent where $P1,\ P2,\ P3$ are stochastic input heat source model parameters defined in Equation 10, $e=\frac{P}{V}$ is the energy density. $Y_{s}$ represents the simulated melt pool dimensions width $\bm{W_{s}}$ or depth $\bm{D_{s}}$.

Here, the HOPGD model is constructed using a set of sampling data of $\bm{W_{s}}$ and $\bm{D_{s}}$ from deterministic AM simulations for different samples of $P1,\ P2,\ P3$ in a predefined parameter space. Once it is known, the stochastic output for $\bm{W_{s}}$ and $\bm{D_{s}}$ can be obtained by giving random input of $P1,\ P2,\ P3$. This procedure is similar to a surrogate model based Monte-Carlo approach for uncertainty propagation. Similar to experimental data, the HOPGD random output distribution is estimated using KDE, and denoted by $f_{Ws}$ and $f_{Ds}$.

Using the above HOPGD model, the optimization problem for finding the appropriate hyper-parameters $\boldsymbol{p^\ast}=[{\mu_1,\mu_2,\mu_3,C}_{11},C_{22},C_{33},C_{12},C_{23},C_{13}]$ can be written as

\begin{equation}
\boldsymbol{p^\ast}=arg\ min\left[\ J(\bm{W_{s}},\bm{W_{e}},{\bm{P}})+J(\bm{D_{s}},\bm{D_{e}},{\bm{P}})\ \right]
\end{equation}

\noindent where $\bm{W_{e}}$ and $\bm{D_{e}}$ are statistical experimental measurements with mean and variance. To define the distance $J$ between experimental and simulated melt pool geometry distributions, the Kullback-Leibler Divergence (KLD) \citep{10.1214/aoms/1177729694} is used. Alternatively, other statistical tests \citep{guo2022identification} approaches can be considered to determine the discrepancy between the experimental and simulated melt pool geometry distributions. The KLD is a measure of the gap between two distributions, and its lowest value indicates the optimal outcome of probability density estimation. A brief description of the KLD can be found in \ref{app3}. The objective function in Eq.\ref{eq16} can further be defined with KLD: 

 \begin{equation}
 \begin{aligned}
{p}^\ast=arg\min
\sum_{i=1}^{11}{f_{We\left(i\right)}\left(W\right)\log{\frac{f_{We\left(i\right)}\left(W\right)}{f_{W_{s}\left(i\right)}\left(P1,P2,P3\right)}}}  +\\ \sum_{i=1}^{11}{f_{De\left(i\right)}\left(D\right)\log{\frac{f_{De\left(i\right)}\left(D\right)}{f_{D_{s}\left(i\right)}\left(P1,P2,P3\right)}}} \label{eq16}
\end{aligned} 
\end{equation}

\noindent where $i$ is the index of single-track cases. $f_{We}, f_{De}, f_{Ws}, f_{Ds}$ are distributions of experimental width, experimental depth, simulated width and depth that can all be calculated from the KDE method discussed above. 

The steps to solve the optimization problem in Eq.\ref{eq16} are as follows:

\begin{enumerate}
    \item 	Sample the parameter space with the adaptive sparse grid strategy \citep{lu2019datadriven,saha2021microscale} and compute the simulated melt pool dimensions $(\bm{W_{s}}, \bm{D_{s}})$ with the AM-CFD model for the selected data points. 
    \item Apply the kernel density estimation (KDE) to the experimental data to obtain the melt pool width and depth distributions $f_{We}({W})$ and $f_{De}({D})$.
    \item Construct HOPGD surrogate model and compute $\bm{W_{s}}$ and $\bm{D_{s}}$ with Eq.\ref{eq13} for sample data.
    \item Generate samples of heat source parameters $P1,\ P2,\ P3$, and obtain stochastic outputs $f_{W_{s}}$ and $f_{D_{s}}$  based on HOPGD surrogate model.
    \item   Solve the optimization problem in Eq.\ref{eq16} with KLD method to calibrate the random heat source parameters and find the optimal hyper-parameters of the stochastic models.
\end{enumerate}

\subsubsection{Validation for single track}
The proposed stochastic AM simulation model provides the capability to predict the variability of the LPBF melt pool using stochastic process parameters. To forecast the stochastic single-track melt pool, we employed the Markov Chain Monte Carlo (MCMC) algorithm \citep{Hastings1970MonteCS}, which is used for sampling from probability distributions based on time series data. MCMC is utilized to generate samples and conduct statistical simulations to predict the relationships between process, structure, and properties.

\begin{figure}[h!]
\centering
\includegraphics[width=1\textwidth]{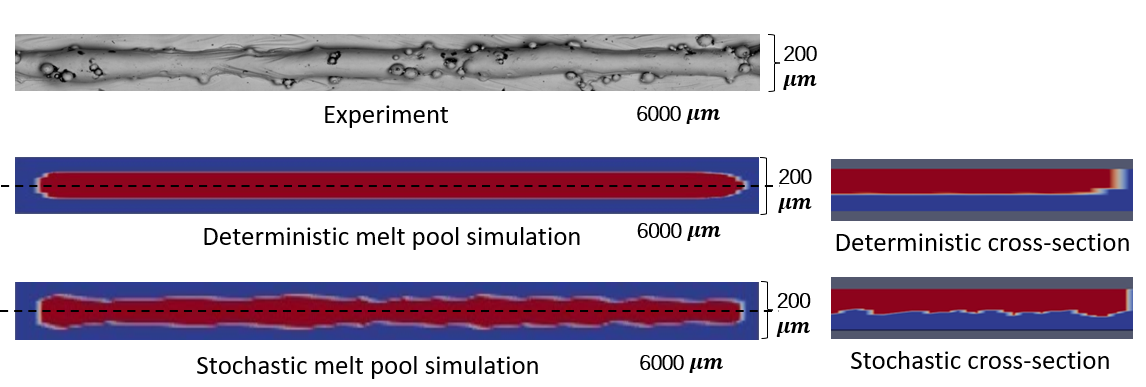}
\caption{ Comparisons between AFRL experiment \citep{Cox2021AFRLAM}, deterministic simulation (with constant heat source model), and stochastic simulation (with calibrated stochastic heat source model) for single track, and cross-section view comparisons between deterministic simulation and stochastic simulation for single track}
\label{fig:validation_single_track}
\end{figure}

\begin{figure}[h!]
\centering
\includegraphics[width=0.8\textwidth]{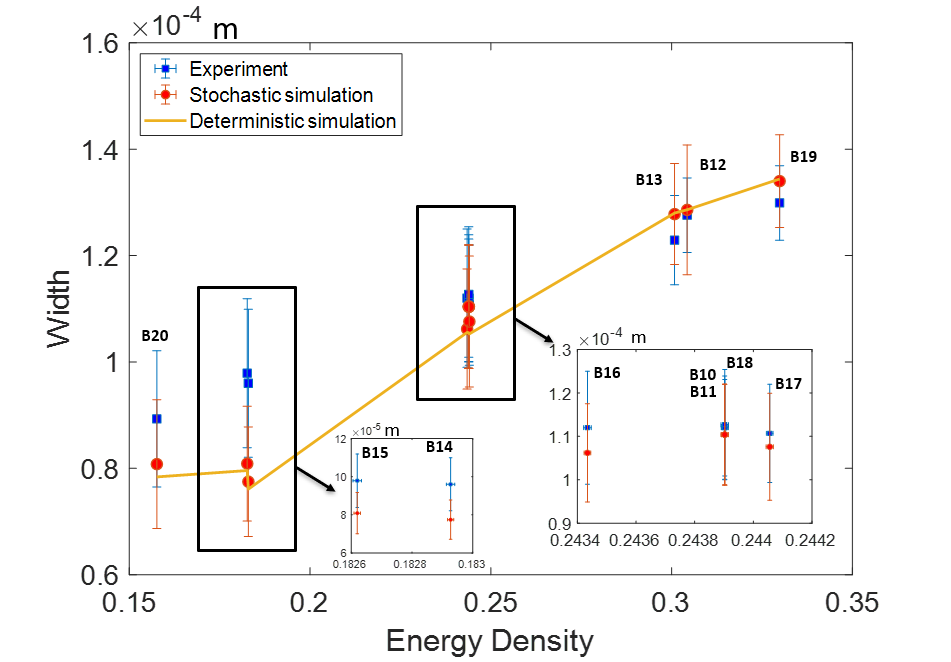}
\caption{Statistical information of melt pool width between stochastic simulation, deterministic simulation and experiment}
\label{fig:width_validation}
\end{figure}

\begin{figure}[h!]
\centering
\includegraphics[width=0.8\textwidth]{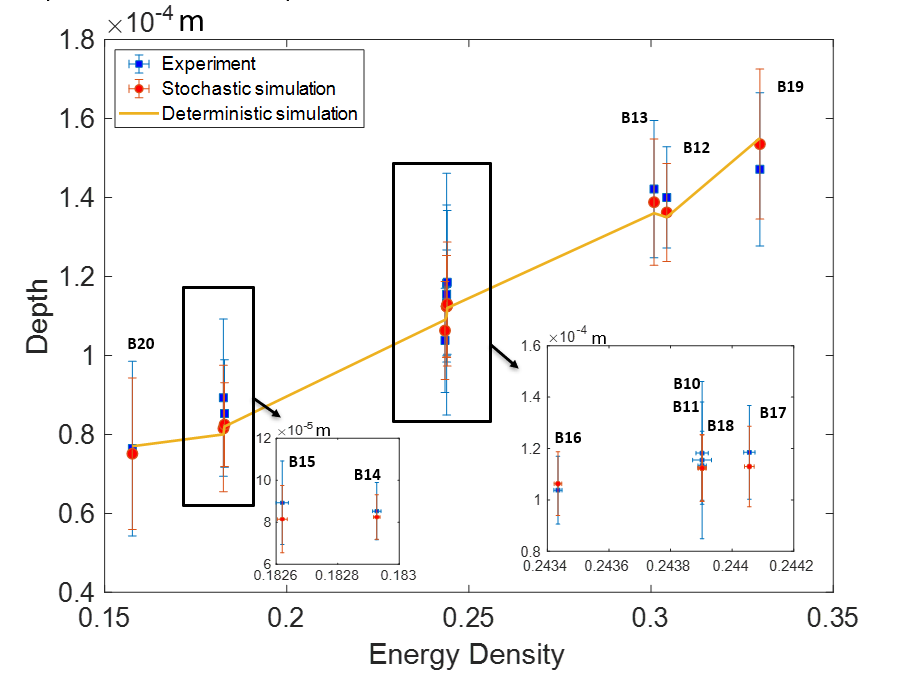}
\caption{Statistical information of melt pool depth between stochastic simulation, deterministic simulation and experiment}
\label{fig:depth_validation}
\end{figure}

In each time step of the AM process simulation, MCMC-sampled heat source parameters are imported into AM-CFD program. This approach enables prediction of surface roughness and porosity for part scale simulations at very reduced computation costs while maintaining a high-fidelity computational model. The comparisons between experiment, deterministic simulation (with constant heat source model), and stochastic simulation (with calibrated stochastic heat source model) is illustrated in Figure \ref{fig:validation_single_track}. The cross-section views in the figure shows the variation in the melt pool in stochastic simulations while the deterministic simulations are unable to capture such uncertain information. To asses the accuracy of the stochastic simulations, statistical melt pool geometry, including mean and variance of the width and depth, is compared with experimental observations \citep{Cox2021AFRLAM} as shown in Figures \ref{fig:width_validation} and \ref{fig:depth_validation}. The figures show energy density for 11 different single-track cases are shown in x-coordinates, while y-coordinates present the melt pool width and depth, respectively. The blue and red error bars represent the mean and variance of the experimental and stochastic simulations, respectively. The yellow line in the figure represents the melt pool dimensions obtained from deterministic simulations. For cases with similar energy density, a zoomed-in  view is provided for detailed comparison. To calculate the mean and variance of melt pool dimensions, 50 width locations and 20 depth locations are measured from both the experimental data and stochastic simulations. The stochastic simulations closely match the experimental melt pool dimensions. In comparison to  deterministic simulation, the stochastic simulation captures the uncertainty in melt pool geometry and provides more accurate predictions for most cases. 

\subsection{Parameterized physics-based machine learning digital twin model (PPB-ML-DT) model}
The choice of a machine learning model over a physics-based simulation model in laser powder bed fusion control is primarily driven by the need for enhanced computational speed. In a control setting, where real-time predictions are essential for making rapid adjustments, traditional physics-based simulations can be prohibitively slow and computationally intensive. Machine learning models, on the other hand, can quickly provide predictions by learning patterns and relationships directly from data, allowing for significantly faster inference times. This speed advantage enables actuators to respond promptly to dynamic changes in the melt pool during the additive manufacturing process, ensuring precise control and optimization of parameters in real-time. To develop a rigorous machine learning model, we take advantage of our calibrated and validated PPB-ML-DT model to generate data as well as the experimental data available by building a robust database for continuous training and update.

Among different machine learning tools, we utilize the Deep AutoRegressive Networks (DARNs) in this context that offers distinct benefits. DARNs excel because they can capture not only the current state of the melt pool, represented by its width and depth, but also the short-term temporal dependencies within the data. Melt pool characteristics in laser powder bed fusion often depend not just on the present input values but also on recent historical information. DARNs are designed to handle such temporal dependencies effectively, making them well-suited for predicting and controlling melt pool parameters. By considering local information and recent trends, these networks facilitate more accurate and responsive control, ensuring the optimal management of critical heat parameters like normalized energy density, heat source radius, and heat source depth in the additive manufacturing process.

The dimensions of the melt pool, along with the associated heat source parameters, serve as the datasets employed to train and evaluate the DARNs. This dataset encompasses paired observations, where the melt pool's width and depth constitute the input, while the output comprises heat source parameters, specifically normalized energy density (NED), heat source depth, and heat source radius. Seventy percent of the data is allocated for training, while 20 percent and 10 percent are designated for validation and testing, respectively. To ensure the network's effective training, we preprocess the data by applying normalization techniques to standardize the input and output data within a consistent range, thereby preventing issues related to gradient problems during training. NED is a dimensionless numerical value that relates processing parameters and material properties, such as material density, heat capacity, and liquidus temperature. It can be expressed as follows \citep{xie2021mechanistic}:

\begin{equation}
    NED = \frac{\eta P}{VHL} \frac{1}{\rho C_p (T_l - T_0)}
\end{equation}

\noindent where $\eta$ stands for absorptivity, $P$ represents laser power, $V$ is the scan speed, and $H$ denotes the hatch spacing. $L$ is the layer thickness, $\rho$ stands for material density, and $C_p$ represents the specific heat. Lastly, $T_l - T_0$ signifies the temperature difference between the melting point and the ambient temperature.

\begin{figure}[h!]
\centering
\includegraphics[width=0.95\textwidth]{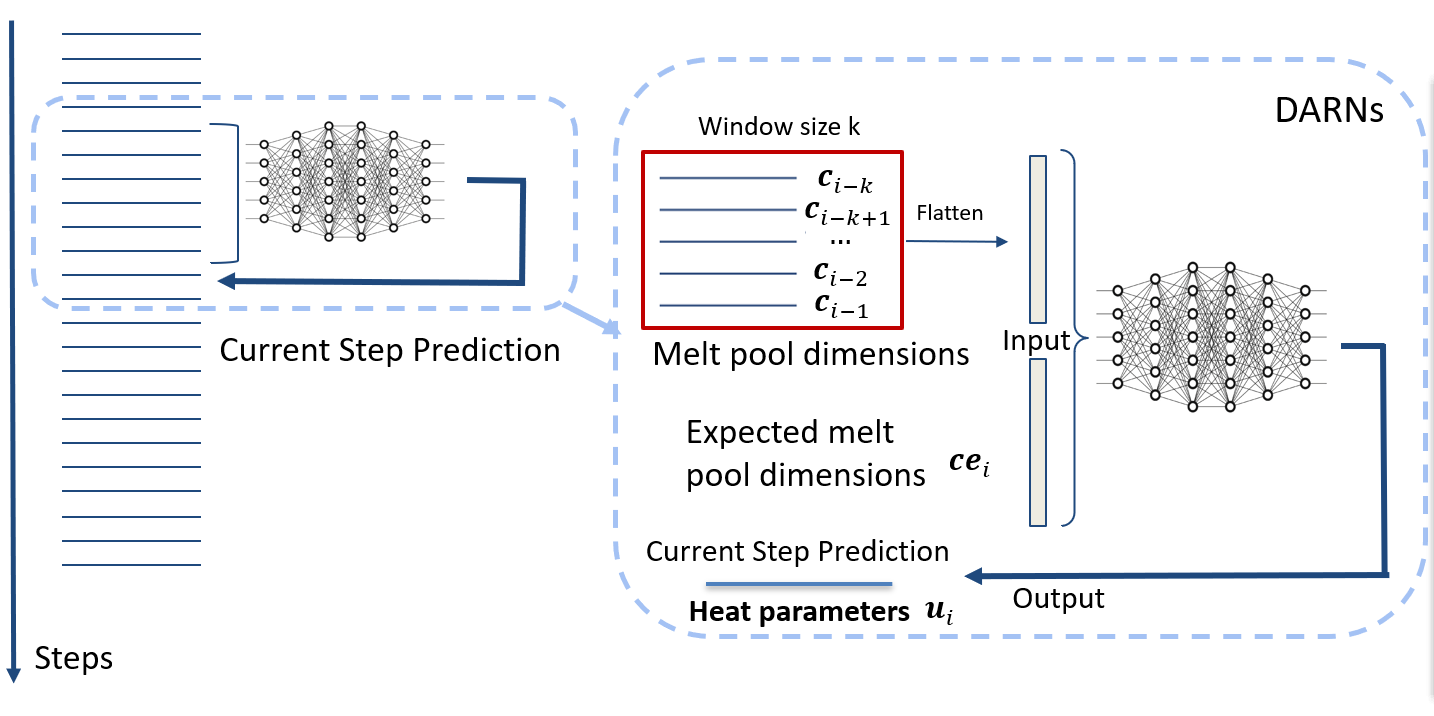}
\caption{The DARNs framework is designed to manage the melt pool dimensions. It takes as input the historical dimensions of the melt pool from previous time steps (denoted as \( \boldsymbol{c_{i-1}}, \boldsymbol{c_{i-2}}, \ldots, \boldsymbol{c_{i-k+1}}, \boldsymbol{c_{i-k}} \)) and the expected dimensions at the current moment (\( \boldsymbol{ce_i} \)). The system outputs the forecasted heating parameters (\( \boldsymbol{u_i} \)) for the current time step. The objective of DARNs is to derive a function \( f_{\theta}(\cdot) \) that provides a close match between the predicted heat parameters (\( \boldsymbol{u_i} \)) and the actual observed values (\( \boldsymbol{x_i} \)).}
\label{fig:DARN}
\end{figure}

The trained DARNs serve as a PPB-ML-DT model which can solve the inverse problem of process control by controlling the process parameters for melt pool geomtry (depth and width). The autoregressive features of DARNs consider the previous 'k' steps as input for each training instance in order to capture the temporal dependencies inherent in melt pool dynamics, where 'k' is defined as the window size. Consider a sequence $\bm{X}=(\bm{x_{0}}, \bm{x_1}, \bm{x_2}, ..., \bm{x_T})$ where $\bm{x_i}$ denotes the heat source input during AM process (NED, heat source depth and heat source radius). We formulate the model in the following way:

The diagram of the DARN is shown in Figure \ref{fig:DARN}. Suppose we have a window size $k$. At time step $i$, we only have access to the historical melt pool dimensions observations such as melt pool width and depth ${\bm{c_{i-1}}, \bm{c_{i-2}}, \ldots, \bm{c_{i-k}}}$ and the expected width and depth at current time step, denoted as $\bm{ce_i}$. Our goal is to establish a function, $f_{\theta}$, such that $\bm{u_i} = f_{\theta}(\bm{c_{i-1}}, \bm{c_{i-2}}, \ldots, \bm{c_{i-k}}; \bm{ce_i})$, which provides the most accurate prediction $\bm{u_i}$ for time step $i$, aiming to closely match $\bm{x_i}$ as much as possible. 

$f_\theta(\cdot)$ can be any parameterized neural network function defined over $\bm{\theta}$. Specifically, we use FFNN in our examples. FFNN is designed to extract features layer-by-layer as defined in the following equations. We utilize $h$ hidden layer FFNN (for our model h = 3) where
\begin{align}
\bm{a^{(1)}} = ReLU(\bm{W_{0}} * \text{input} + \bm{B_{0}})\\
\bm{a^{(l)}} = ReLU(\bm{W_{l-1}} * \bm{a^{(l-1)}} + \bm{B_{l-1}}) \\
\text{output} = Sigmod(\bm{W_{h-1}} * \bm{a^{(h-1)}} + \bm{B_{h-1}})
\end{align}

Note that the learnable parameters in this networks $\bm{\theta}=\{\bm{W_{0}}, \bm{B_{0}}, ..., \bm{W_{h}}, \bm{B_{h}}\}$. $\bm{W_{i}}$ and $\bm{B_{i}}$ are the weights and bias of the i-th hidden layer.

In order to train $f_\theta(\cdot)$, we utilize maximum likelihood estimation to construct the loss function and solve the following optimization problem:

\begin{equation}
  min_{\bm{\theta}} \ \sum_i\text{Log}(\text{P}_{\bm{x_i}}(\bm{u_i} | \bm{c_{i-1}}, \bm{c_{i-2}}, ..., \bm{c_{i-k}}; \bm{ce_i} ; \bm{\theta}))  
\end{equation}

The conditional distribution of $\bm{u_i}$ achieve the maximum likelihood when $\bm{u_i} = \bm{x_i}$, i.e. the prediction equals the real experimental results. During the optimization process, the likelihood of $\bm{u_i}$ increase gradually until the likelihood is maximized over all training data. In practices, we assume $P_{\bm{x_i}} \sim N(\bm{x_i}, \bm{I})$ follows a Gaussian distribution centered at $\bm{x_i}$ with identity covariance matrix, therefore, the optimization problem becomes to the following:

\begin{align}
&min_{\bm{\theta}} \ \sum_i\text{Log}(\text{P}_{\bm{x_i}}(\bm{u_i} | \bm{c_{i-1}}, \bm{c_{i-2}}, ..., \bm{c_{i-k}}; \bm{ce_i} ; \bm{\theta}))\\
&min_{\bm{\theta}} \ \sum_i\text{Log}(\frac{1}{2\pi} \text{Exp} (-\frac{1}{2} ||\bm{x_i} - f_{\bm{\theta}}(\bm{c_{i-1}}, \bm{c_{i-2}}, ..., \bm{c_{i-k}}; \bm{ce_i}; \bm{\theta})||^2))\\
&min_{\bm{\theta}} \ \sum_i^n||\bm{x_i} -  f_{\bm{\theta}}(\bm{c_{i-1}}, \bm{c_{i-2}}, ..., \bm{c_{i-k}}; \bm{ce_i}; \bm{\theta})||^2) 
\end{align}

The Adam Optimizer is employed for the optimization of the parameter \(\bm{\theta}\), as suggested in the work by Kingma and Ba \citep{kingma2014adam}. The network was trained using an initial learning rate of \(1 \times 10^{-4}\) across 2450 epochs. Batch training was adopted to providing sufficient update frequency for stable convergence and robust generalization during training, utilizing a batch size of 64.

\section{Statistical predictions and diagnostics applications of PPB-DT model}
\label{sec3}
\subsection{Statistical predictions applications of PPB-DT model}
The PPB-DT model is a stochastically calibrated and validated physics-based model with parameterized heat source model. The model is used to simulate multi-layer (thin-wall) and multi-track cases and validate against the AFRL experiment for surface roughness and LOF porosity. 

\subsubsection{Predictions of surface roughness of thin-wall samples}
\begin{figure}[h!]
\centering
\includegraphics[width=1\textwidth]{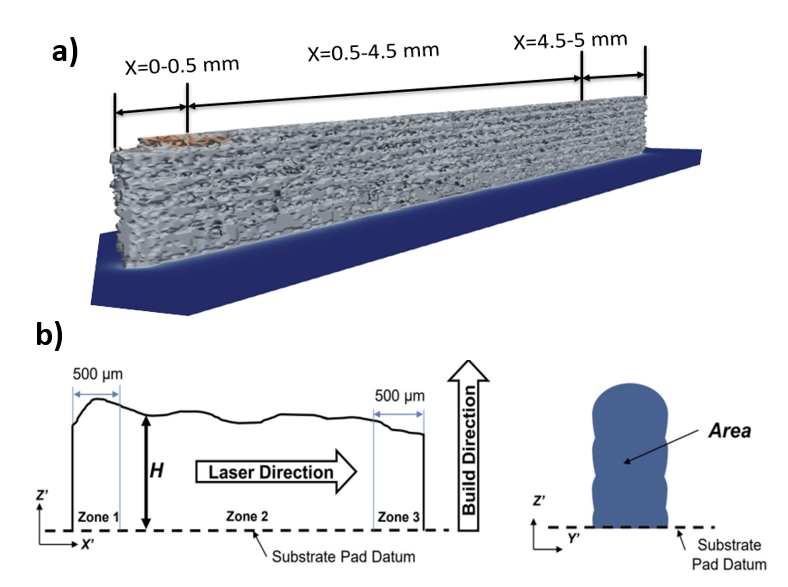}
\caption{As-built multi-layer structure and its measurements for case B1. a) Multi-layer simulation. b) A schematic of the height and cross section area measurements for three Zones \citep{Cox2021AFRLAM}}
\label{fig:multilayer}
\end{figure}

\begin{figure}[h!]
\centering
\includegraphics[width=1\textwidth]{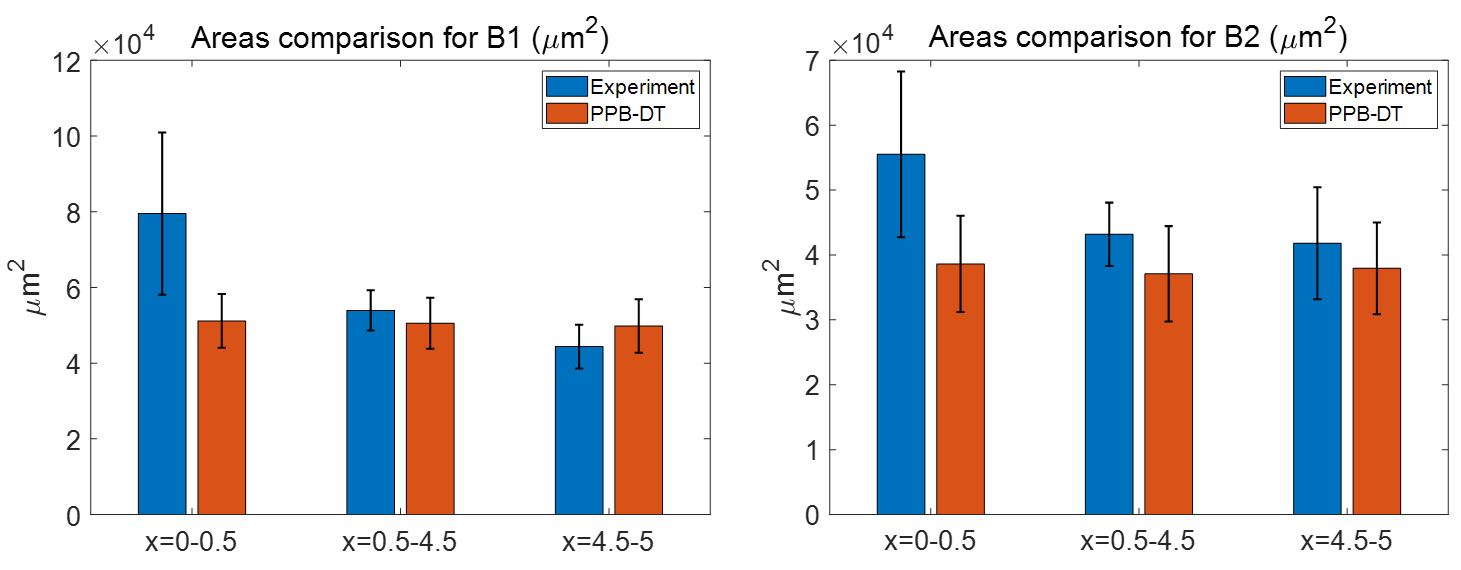}
\caption{Quantitative comparisons of cross-sectional area between experimental measurements and PPB-DT predictions}
\label{fig:multilayer_compare}
\end{figure}

Two thin-wall specimens, B1 and B2, are simulated, each consisting of 10 consecutive layers with a thickness of 40 $\mu m$ and unidirectional scanning track length of 5 mm \citep{Cox2021AFRLAM}. The process parameters for these multilayer specimen are summarized in Table \ref{tab:tab4}. Specimen B1 used a laser power of 300 W and scan speed of 1230 mm/s, and specimen B2 used 241 W and 1529 mm/s. Figure \ref{fig:multilayer}a shows the simulated result for case B1. For quantitative comparison, the wall is divided into three measurement zones depending on positions (see Figure \ref{fig:multilayer}). The average height (mean and standard deviation) above the substrate pad datum and the total cross-sectional area for the entire portion of the wall above the substrate pad datum were measured for each zone as shown in Figure \ref{fig:multilayer}b. Three cross sections are collected within Zones 1 and 3, while approximately 20 cross sections are collected in Zone 2. Figure \ref{fig:multilayer_compare} presents the comparisons of the three cross sectional area for three different zones between the experimentally measured and computationally predicted values for B1 and B2 multi-track cases. The simulated height and area closely match with the measurements in the second and third zones, indicating the developed model can accurately predict the steady-state melt pool geometry. However, in Zone 1, the beginning region of each layer, the model underestimates the melt pool geometry. This suggests that some transient behaviors occurring at the beginning of each layer are not adequately captured by the model.

\begin{table}[]
\caption{Multi-layer simulation process parameters \citep{Cox2021AFRLAM}}
\label{tab:tab4}
\resizebox{\textwidth}{12mm}{
\begin{tabular}{|c|c|c|c|c|c|}
\hline
\begin{tabular}[c]{@{}c@{}}Case \\ Number\end{tabular} & \begin{tabular}[c]{@{}c@{}}Laser \\ Power (W)\end{tabular} & \begin{tabular}[c]{@{}c@{}}Scan \\ Speed (mm/s)\end{tabular} & \begin{tabular}[c]{@{}c@{}}Layer \\ thickness ($\mu$ m)\end{tabular} & \begin{tabular}[c]{@{}c@{}}Track \\ length (mm)\end{tabular} & \begin{tabular}[c]{@{}c@{}}The number \\ of layers\end{tabular} \\ \hline
B1                                                    & 300                                                        & 1230                                                         & 40                                                                               & 5                                                            & 10                                                              \\ \hline
B2                                                    & 241                                                        & 1529                                                         & 40                                                                               & 5                                                            & 10                                                              \\ \hline
\end{tabular}
}
\end{table}

\begin{figure}[h!]
\centering
\includegraphics[width=1\textwidth]{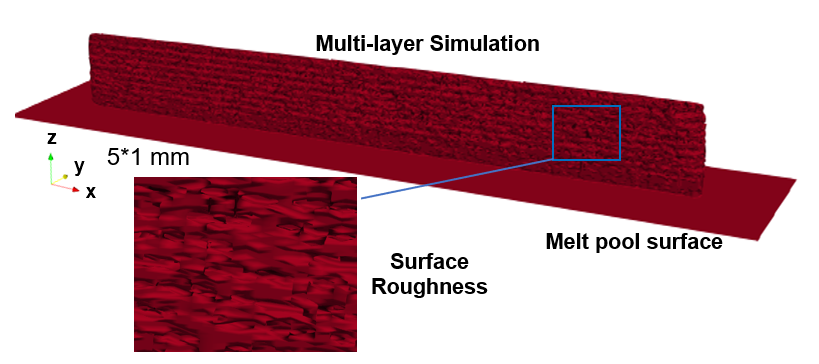}
\caption{Surface roughness prediction for multi-layer simulation}
\label{fig:b1_roughness}
\end{figure}

Further, the multilayer case B1 (see Table \ref{tab:tab4}) is simulated using the PPB-ML-DT model to predict the surface roughness and compare with the experiment. The laser power for this case was $300 W$ and scan speed was $1230 mm/s$ which is a combination of high power and low scanning speed for the Inconel 625 alloy. Figure \ref{fig:b1_roughness} presents the PPB-DT predictions of the multilayer case B1, which manifests the surface roughness due to the stochastic AM process.

The primary roughness parameter reported is the arithmetic mean height ($Sa$) that evaluates the average standard deviation of the heights from the mean plane (valleys and peaks) in a surface profile to compute the degree of roughness. To compute $Sa$, first the fitting plane for the points collected from the surface are calculated. Then, the height of a peak or valley is determined by evaluating the height coordinate of each point in the dataset. 

\begin{figure}[h!]
\centering
\includegraphics[width=0.8\textwidth]{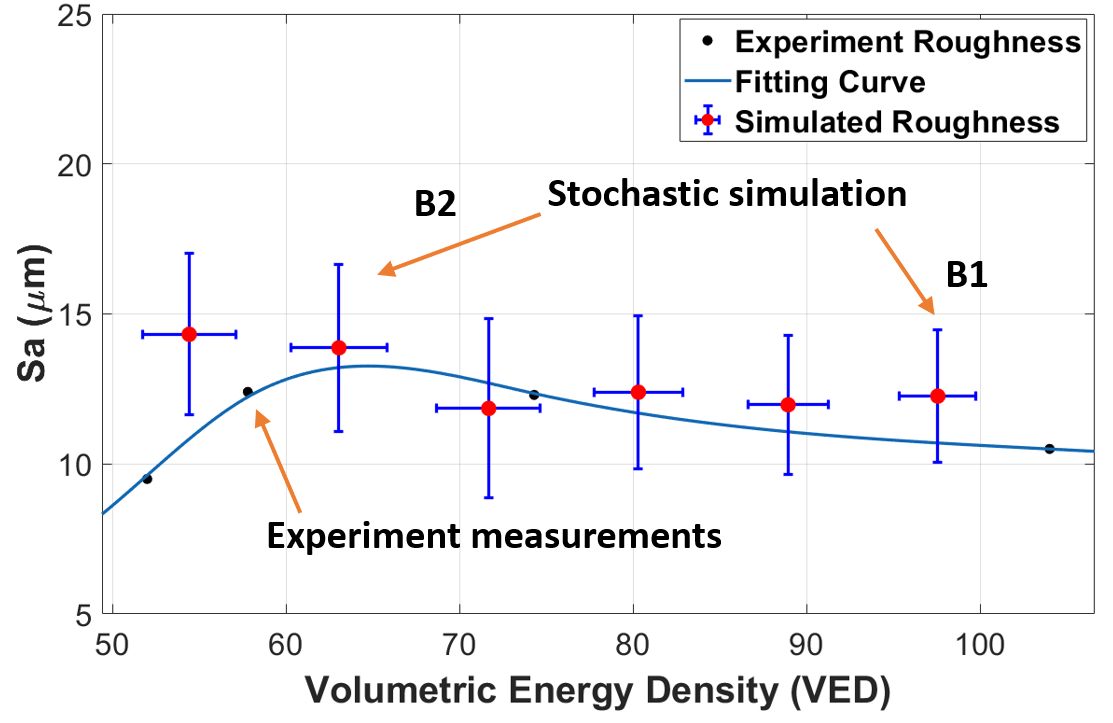}
\caption{Validation between simulated surface roughness and experimental data (Imade Koutiri et.al., 2018) under different VEDs. The blue dots are experiment measurements, the blue line is the fitting curing of experiment points. The error bars with red center represent the statistical measurement of simulated surface roughness, in which B1 and B2 have the same VED as the AFRL experiment.}
\label{fig:roughness_compare}
\end{figure}

The equation of average roughness $Sa$ is given by \citep{degarmo1997materials}:
 \begin{equation}
Sa=\frac{1}{A}\iint_{S}\left|f(y)\right|dS
\end{equation}
\noindent where $A$ is the sampling area and $f(y)$ is height of the profile. The simulated wall is equally divided into 10 regions, and the mean value and variance for the surface roughness is calculated. The calculated surface roughness for case  B1 and B2 is $12.62 \pm 2.61 \mu m$,  and $Sa=14.57 \pm 3.18 \mu m$, respectively. 
To understand the effect of processing conditions on surface roughness, we plot the surface roughness against the volumetric energy density (VED) \citep{bertoli2017limitationsr}, which is defined as:

\begin{equation}
VED=\frac{P}{V \sigma_b t}
\end{equation}
\noindent where $P$ is laser power, $V$ is scan speed, $\sigma_b$ is the laser beam diameter, and $t$ is the thickness for a single layer. VEDs for B1 and B2 multi-layer cases are $ 97.56 J/{mm}^3$ and $ 63.05 J/{mm}^3$, respectively.
The predicted surface roughness for AFRL case B1 and B2 is shown (see Figure \ref{fig:roughness_compare}) against volumetric energy density with statistical limit.  For the same volumetric energy density, the surface roughness is also compared with experiments \citep{koutiri2018influence} on same material Inconel 625, represented by the black dots in the figure. A blue curve is drawn by fitting the experimental data of surface roughness. The two error bars represent the simulated surface roughness with corresponding mean and variance. The comparison indicates that the simulated surface roughness matches well with the experiment, thereby validating the accuracy of our PPB-ML-DT model. 

\begin{table}[]
\caption{Multi-track simulation process parameters \citep{Cox2021AFRLAM}}
\label{tab:tab5}
\resizebox{\textwidth}{20mm}{
\begin{tabular}{|c|c|c|c|c|c|}
\hline
\begin{tabular}[c]{@{}c@{}}Case \\ Number\end{tabular} & \begin{tabular}[c]{@{}c@{}}Laser \\ Power (W)\end{tabular} & \begin{tabular}[c]{@{}c@{}}Scan \\ Speed (mm/s)\end{tabular} & \begin{tabular}[c]{@{}c@{}}Hatch \\ Spacing (mm)\end{tabular} & \begin{tabular}[c]{@{}c@{}}Toolpath plane \\ dimensions (mm)\end{tabular} & \begin{tabular}[c]{@{}c@{}}The number \\ of tracks\end{tabular} \\ \hline
C1                                                    & 300                                                        & 1230                                                         & 0.1                                                           & 3*3                                                                       & 30                                                              \\ \hline
C2                                                    & 300                                                        & 1230                                                         & 0.1                                                           & 10*3                                                                      & 30                                                              \\ \hline
C3                                                    & 300                                                        & 1230                                                         & 0.075                                                         & 10*3                                                                      & 40                                                              \\ \hline
C4                                                    & 300                                                        & 1230                                                         & 0.125                                                         & 10*3                                                                      & 24                                                              \\ \hline
C5                                                    & 300                                                        & 1230                                                         & 0.1                                                           & 10*3                                                                      & 30                                                              \\ \hline
C6                                                    & 290                                                        & 953                                                          & 0.1                                                           & 15*3                                                                      & 30                                                              \\ \hline
\end{tabular}
}
\end{table}

\begin{figure}[h!]
\centering
\includegraphics[width=0.8\textwidth]{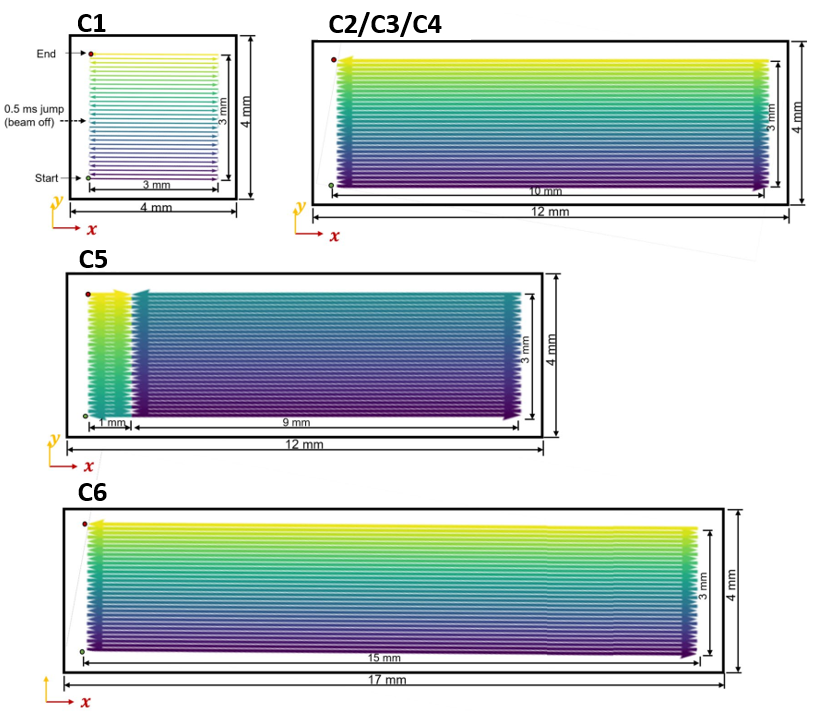}
\caption{Scan strategies for the multi-track cases \citep{Cox2021AFRLAM}}
\label{fig:multi-track_scan}
\end{figure}

\begin{figure}[h!]
\centering
\includegraphics[width=0.9\textwidth]{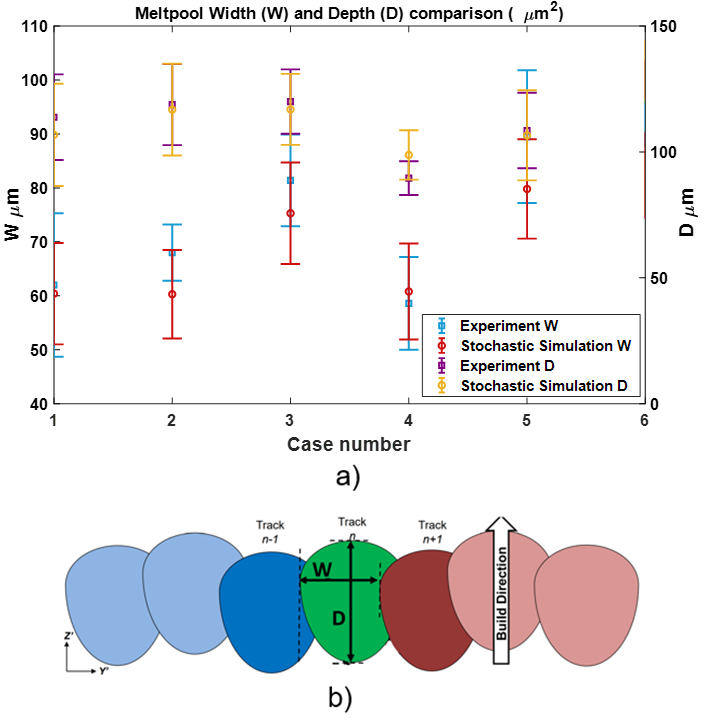}
\caption{Quantitative comparisons of melt pool dimensions width ($W$) and depth ($D$) between experimental and stochastic simulations at the middle of the toolpath (x=1.5mm) for six multi-track cases. The average and standard deviation of each quantity for different tracks are plotted (The error bar represents the standard deviation.) x label "Case number" stands for six multi-track cases. }
\label{fig:multi-track_compare}
\end{figure}

\subsubsection{Predictions of melt pool geometries of multi-track samples}

Six simulations of multi-track cases were conducted using a calibrated stochastic AM-CFD model to predict the geometrical characteristic of the melted multi-track cases in the LPBF process. Figure \ref{fig:multi-track_scan} shows the substrate geometries and tool paths used for these simulations, labeled as C1, C2, C3, C4, C5, and C6, corresponding to LPBF experiments performed by AFRL. A dwell time of 0.5ms was set between the end of scan of one layer to the beginning of the next layer. During this dwell time period, the laser beam was turned off. The black frames show the substrate dimensions, and the arrows represent the laser scan paths. Table \ref{tab:tab5} summarizes the process parameters used for all six multi-track cases. Figure \ref{fig:multi-track_compare} a shows the quantitative comparisons of melted track geometries at the middle of the toolpath (x=1.5mm) for the six multi-track simulations for the average and standard deviation of the melt pool width ($W$) and depth ($D$). The multi-track simulations closely match with experimental data, and demonstrate potential for high-precision AM predictions. Additionally, these simulations can be used for prediction of surface roughness and LOF porosity at significantly reduced computation costs.

\subsubsection{Predictions of LOF porosity for multilayer multi-track samples}
The PPB-DT model is utilized to predict the surface roughness and LOF porosity for the multilayer multi-track cases of AFRL experiments. Additionally, we demonstrate the prediction of surface roughness and LOF porosity for an as-built part using a multilayer multi-track case. Markov chain Monte Carlo sampling is employed to generate time-dependent sequences of the processing conditions to simulate the part scale.

\begin{figure}[h!]
\centering
\includegraphics[width=1\textwidth]{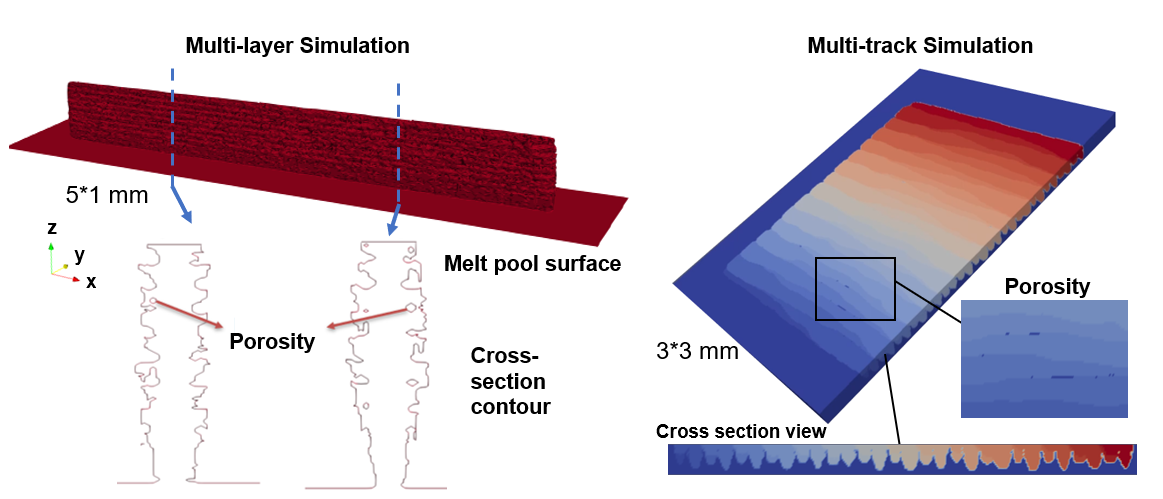}
\caption{Porosity prediction for multi-layer and multi-track simulation }
\label{fig:porosity}
\end{figure} 

For the multilayer and multi-track cases are investigated for the LOF  porosity through our stochastic AM simulation. Figure \ref{fig:porosity}a show the multilayer case B1 (see Table \ref{tab:tab4} where the porosity is visible in between the layer and near the corner of the track. These porosity occurs due to the improper melting of the powder and tracked in our model by tracking the melting temperature of the scan. Figure \ref{fig:porosity}a show the multi-track case, where the LOF porosity is between the consecutive tracks. Also, the multi-track case simulated using stochastic AM model reveals the non-uniform melt pool shape and size distribution which can not be captured in a deterministic model. 

\begin{figure}[h!]
\centering
\includegraphics[width=0.77\textwidth]{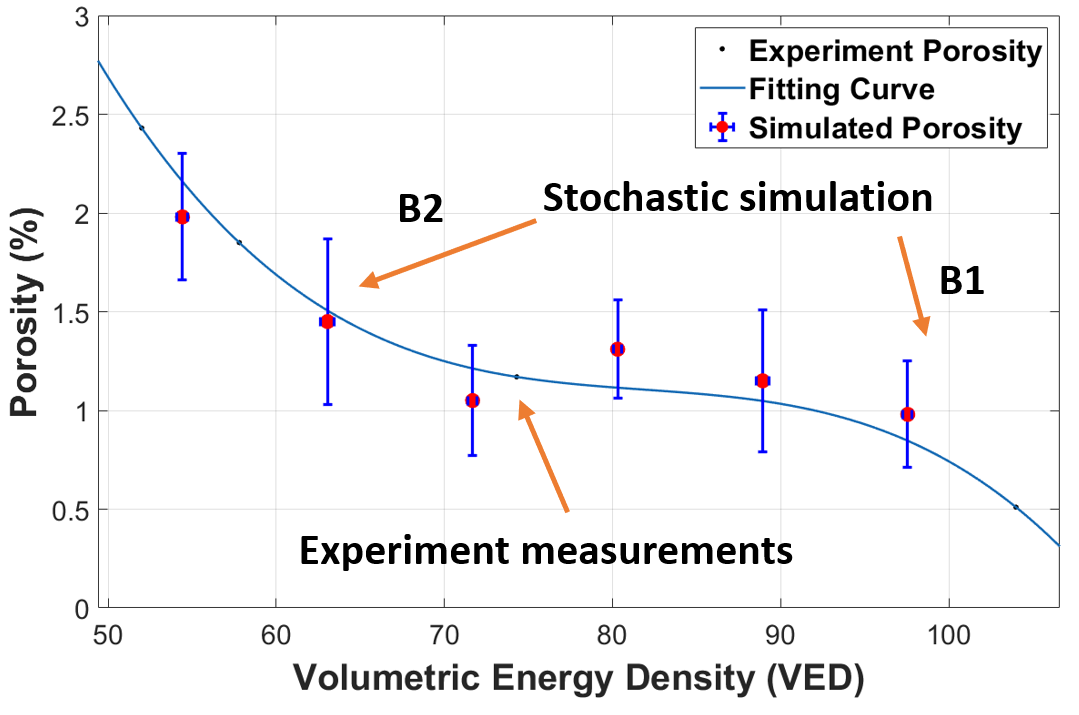}
\caption{Validation between simulated porosity and experimental data (Imade Koutiri et.al., 2018) under different VEDs. The blue dots are experiment measurements, the blue line is the fitting curing of experiment points. The error bars with red center represent the statistical measurement of simulated porosity,in which B1 and B2 have the same VED as the AFRL experiment.}
\label{fig: porosity_compare}
\end{figure}

In Figure \ref{fig: porosity_compare}, the variation of predicted LOF porosity with the volumetric energy density is presented and compared with experimental cases \citep{koutiri2018influence}. The LOF porosity decreases as the volumetric energy increases which means a better powder melting scenario. It should be noted that, experimentally measure porosity includes all mode of porosity; however, for the chosen VED ranges, the LOF porosity is the dominant mode and other mechanism of porosity formation is negligible. 

\subsection{Defects diagnostics applications of PPB-DT model}
In this section, a part scale defects diagnostic application is demonstrated using the PPB-ML-DT model. For the demonstration, National Institute of Standards and Technology (NIST)  overhang part X4 \citep{lane2020process,praniewicz2020x} has been used. Same geometry and scan strategy used by NIST for the part has been used in our simulation; however, we only simulated 1/8 part of the sample due to the expensive thermo-fluid simulation in our PPB-ML-DT model. The goal of this is to show the defect diagonsis capability of the PPB-ML-DT model for the surface roughness and LOF porosity for a part-scale level. 
\begin{figure}[h!]
\centering
\includegraphics[width=0.9\textwidth]{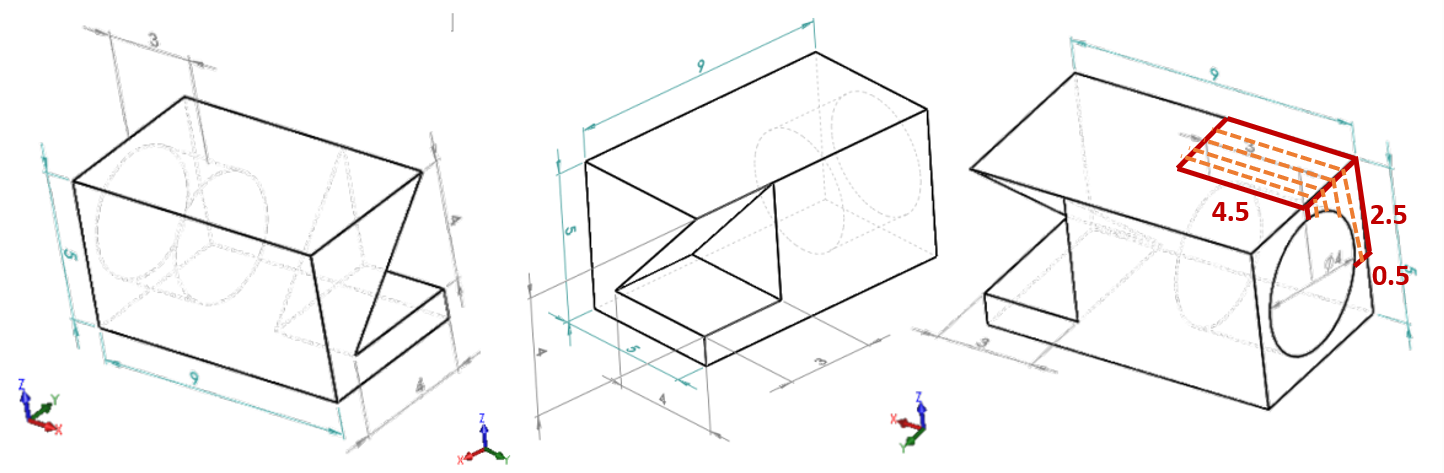}
\caption{Three view orientations geometry of overhang part (unit:mm), the red region defines the simulation part}
\label{fig:Figure17_new2}
\end{figure}

\begin{figure}[h!]
\centering
\includegraphics[width=0.77\textwidth]{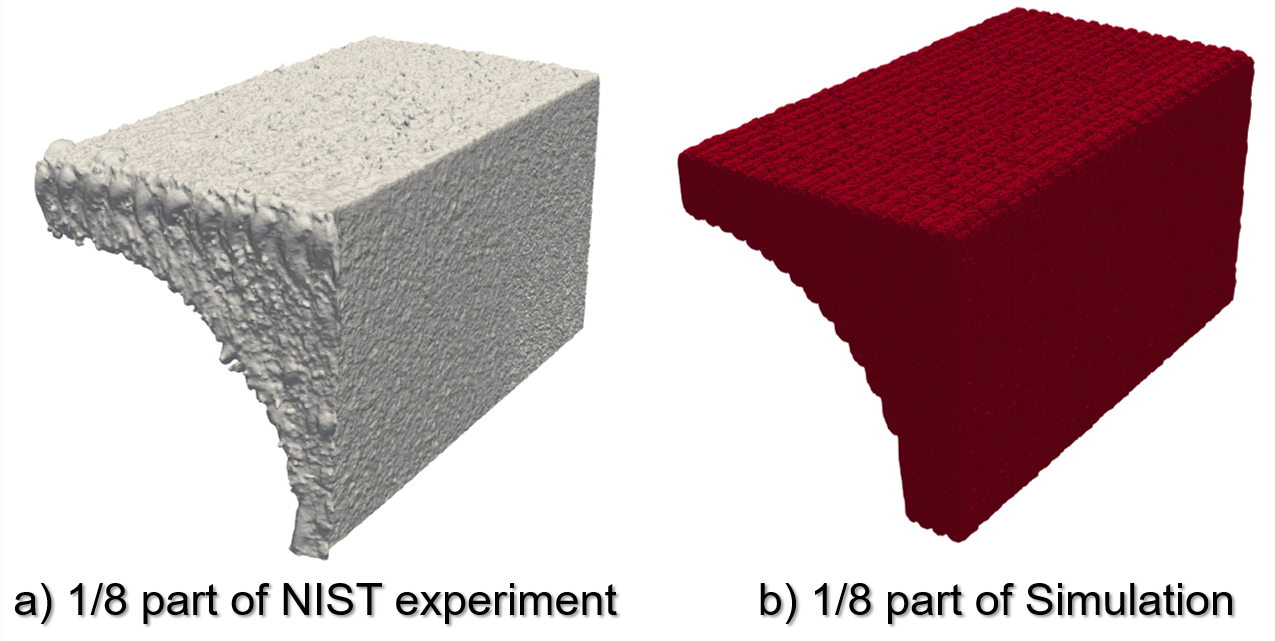}
\caption{Comparisons between experimental a) and simulation b) result of 1/8 part of Overhang Part X4}
\label{fig:overhangresult}
\end{figure}

The NIST “Overhang Part X4” \citep{lane2020process,praniewicz2020x} is fabricated on the Additive Manufacturing Metrology Testbed (AMMT) from nickel superalloy Incolnel 625 (IN625). The part has a $9mm \times 5mm \times 5mm$ rectangular prism shape with a 45° overhang feature and a horizontal cylindrical cutout. Three different view orientations of the computer-aided design (CAD) part geometry is shown in Figure \ref{fig:Figure17_new2} . For demonstration purpose, 1/8 part is simulated and compared with experimental result qualitatively. The dimension of the 1/8 part is 4.5mm x 2.5mm x 2.5mm, as demonstrated by the red region in Figure \ref{fig:Figure17_new2}. To enhance the computational speed, we partitioned the entire 100 layers into distinct groups and executed them through a parallel high-performance computing. 
\begin{table}[]
\caption{Process parameters of NIST AM part}
\label{tab:5}
\begin{tabular}{|c|c|c|c|c|c|}
\hline
P ($W$) & V ($mm/s$) & \begin{tabular}[c]{@{}c@{}}Layer \\ Thickness ($mm$)\end{tabular} & \begin{tabular}[c]{@{}c@{}}Hatch \\ Spacing ($mm$)\end{tabular} & \begin{tabular}[c]{@{}c@{}}Layer \\ number\end{tabular} & \begin{tabular}[c]{@{}c@{}}Volumetric \\ energy density \\ ($J/mm^{3}$)\end{tabular} \\ \hline
300   & 800      & 0.04                                                            & 0.1                                                           & 100                                                      & 101.56                                                               \\ \hline
\end{tabular}
\end{table}

\begin{figure}[h!]
\centering
\includegraphics[width=0.85\textwidth]{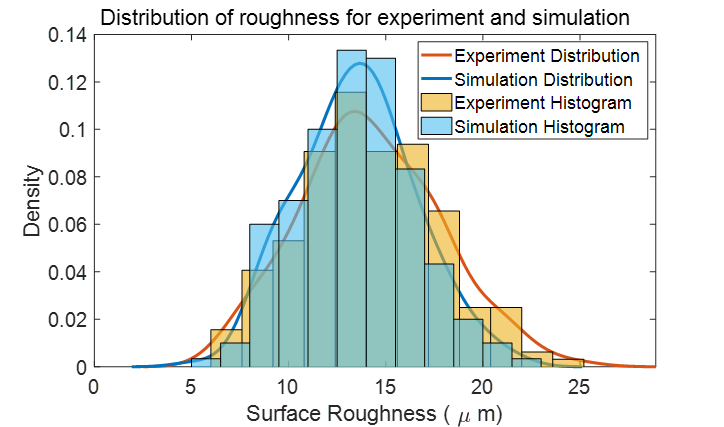}
\caption{Comparisons surface roughness distribution and histogram between experimental and simulation for 1/8 part of Overhang Part X4}
\label{fig:overhang_distribution}
\end{figure}

The process parameters of the overhang part is shown in the Table \ref{tab:5}. Figure \ref{fig:overhangresult} presents the surface finish comparison between the NIST experiment and our simulation. Both the experiment and stochastic simulation reveal an irregular and uneven surface when viewed from the front, as well as a rough surface with linear patterns along the top (aligned with the build direction). The proposed stochastic AM simulation is thereby shown to possess the capability of simulating additive manufacturing (AM) parts with defects.
The part is divided into five regions to compute the surface roughness of the front surface (as shown in the orange dash lines in Figure \ref{fig:Figure17_new2}). The predicted roughness is $Sa_{sim}=13.09 \pm 3.01 \mu m$, and it has been validated against the experimental roughness measurement $Sa_{exp}=14.44 \pm 3.59 \mu m$, with a difference of $9.4$ percent for the mean values. A distribution of the experimental surface roughness is shown with the PPB-DT simulation predicted distribution. Both the distribution matched closley for the mean and the standard deviation (see Figure \ref{fig:overhang_distribution}). 

The PPB-ML-DT model can also predict the LOF porosity the NIST part. The predicted LOF porosity is $0.52 \pm 0.24 \% $. The experimental porosity information is not available for the NIST overhang part.

\section{Online monitoring and control applications of PPB-ML-DT model}
\label{sec4}
In LPBF process melt pool monitoring is a crucial aspects to maintain part quality and certification. By controlling the processing parameters such as laser power, speed, etc., desired melt pool geometries can be achieved which can ensure desired performance. However, this is a challenging inverse problem from the modeling perspective where we need to predict the processing conditions for a desired melt pool phenomena. For a control application, such predictions has to be made in real-time (in milliseconds) to control the process for immediate layer.  The real-time prediction ability of the machine learning tool motivated us to further develop a s machine learning-based digital twin model for control applications. 

In the machine learning model, we set up an inverse problem to predict heat parameters, namely, normalized energy density (NED), heat source radius, and heat source depth for inputs of melt pool width and depth. Specifically, we choose a DARNs model to capture the transient dynamics of the melt pool. 

\begin{figure}[ht!]
    \centering
    \includegraphics[width=1\linewidth]{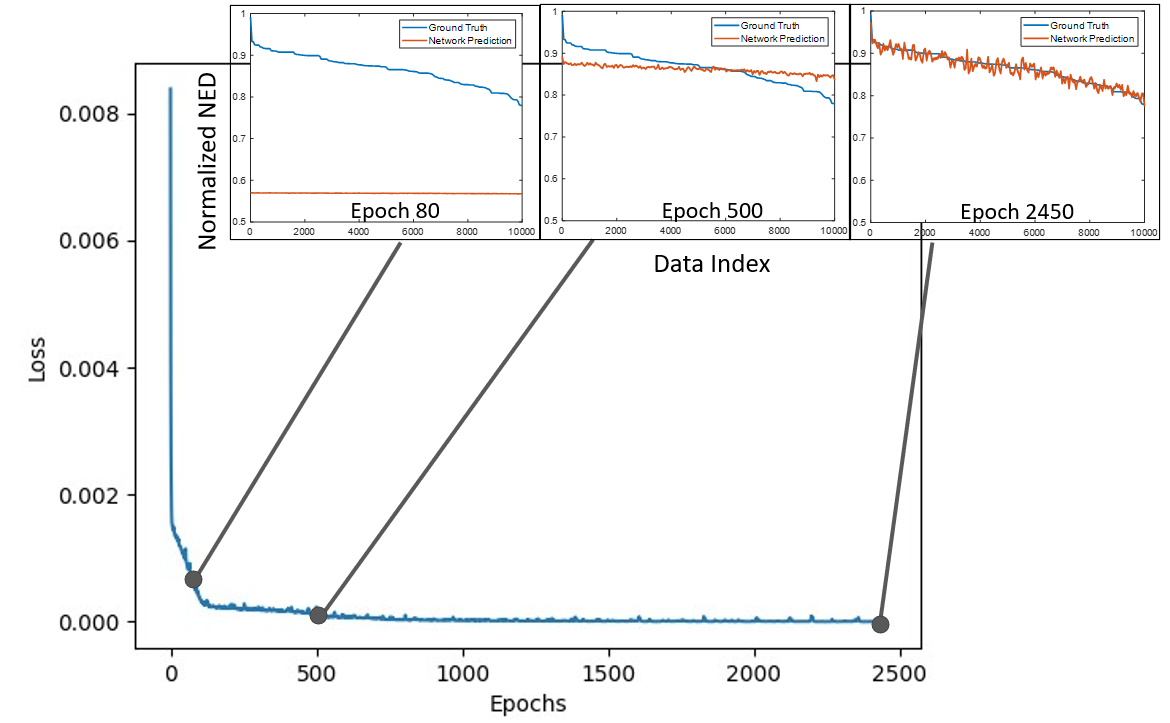}
    \caption{Loss during the optimization epochs. Each epoch denotes a full iteration over all the training data.}
    \label{fig:mllossfigure}
\end{figure}

The figure presented below illustrates the relationship between the number of network epochs and the corresponding loss function of the optimization problem defined in Section 2.3. The loss function quantifies the likelihood across all training data while considering the influence of adjacent data points based on a windowing function. As depicted in Figure \ref{fig:mllossfigure}, the training process of our model demonstrates a notable reduction in loss over the course of the epochs. Initially, from epoch 0 to 120, a substantial decrease in loss is observed, followed by a gradual decline in the loss function. Remarkably, the model exhibits ongoing improvement on the evaluation datasets, persisting until epoch 2450.

Furthermore, Figure \ref{fig:mllossfigure} provides sub-figures depicting a comparative analysis between the ground truth (represented by the blue line) and the network's predictions (illustrated by the orange line) at various epochs (specifically, epochs 80, 500, and 2450). A clear trend emerges wherein the network's predictions increasingly align with the ground truth as the training progresses. It is worth noting that the $x$ index of the subplots for these comparisons have been sorted in order to clearly represent the results.

 Figure \ref{fig:mllossfigure} illustrates the discrepancy between the ground truth and network predictions for the test data. The the error between the ground truth and prediction is quantified using the following equation. It also demonstrates that the errors remain consistently below 5 percent for the test data at epoch 2450, employing a window size of 6. This highlights the model's robust performance in accurately predicting the target values, even when considering variations in the input data. The equation for computing the relative error is as follows:

\begin{equation}
    error=\frac{\|ml\_model-ground\_truth\|}{ground\_truth}
\end{equation}

\noindent where $ml\_model$ is the result obtained from our autoregressive machine learning model, $ground\_truth$  refer to the ground truth data.

\begin{figure}[ht!]
    \centering
    \includegraphics[width=0.9\linewidth]{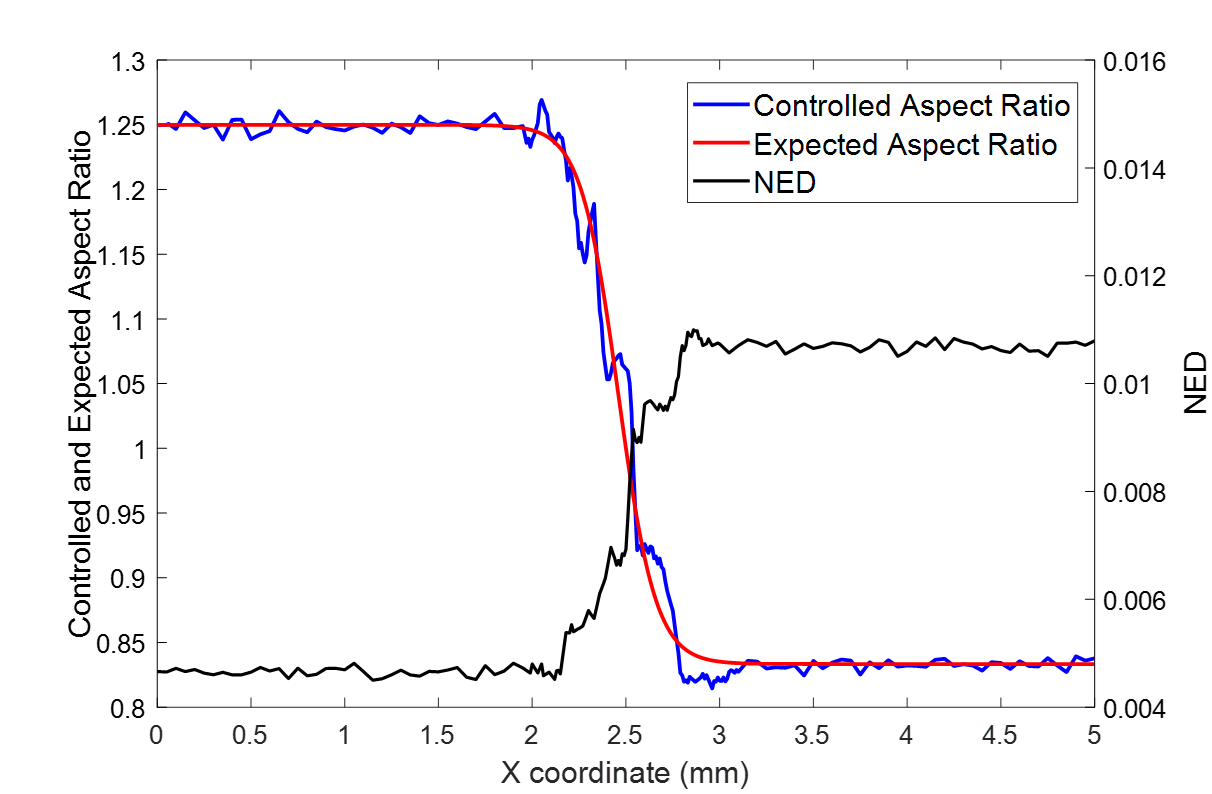}
    \caption{Comparison of the expected melt pool aspect ratio with the controlled one. The expected aspect ratio are traced by red curves, whereas the blue curves depict the simulations controlled by the heat source parameters produced by the DARNs model. The heat source parameter like NED is plotted in black.}
    \label{fig:control}
\end{figure}

\begin{figure}[ht!]
    \centering
    \includegraphics[width=0.85\linewidth]{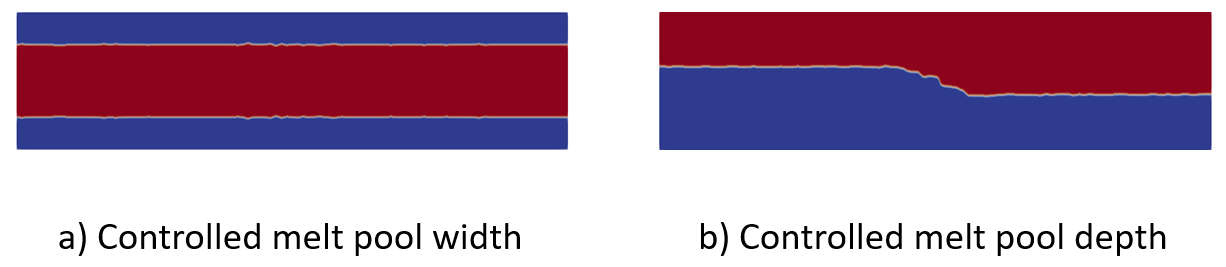}
    \caption{Controlled melt pool a) width and b) depth}
    \label{fig:controlled_dimension}
\end{figure}

To assess the control proficiency of our machine learning model, we present the anticipated melt pool depth, which follows the Sigmoid function profile, and we maintain the expected melt pool width at fixed values. The expected aspect ratio of the melt pool, calculated as the width to depth ratio, is represented by the red line in the Figure \ref{fig:control}. The corresponding heat source parameters were derived from the PPB-ML-DT model. These parameters were then fed into our physics-based AM-CFD solver to obtain melt pool dimensions. Following that, we derived the aspect ratio from the predicted dimensions, which is depicted by the blue line on the graph, and contrasted with our expected values. The heat source parameters generated by the network, such as NED, are illustrated by the black curve in Figure \ref{fig:control}. The controlled melt pool and width and depth are shown in Figure \ref{fig:controlled_dimension}. The Notably, our actual results align closely with the expected melt pool dimensions. The melt pool control is particularly accurate during the steady stages (the initial and final segments of the curve). However, sudden changes in the expected melt pool dimension lead to a slight increase in discrepancies. Most of these deviations remain under 5 percent, with the largest discrepancy being $6.7$ percent.

\section{Discussion}
\label{sec5}
 One of the key advantage of using stochastic AM-CFD model as PPB-DT is that it can capture the transient nature of the melt pool phenomena, which results in better prediction of surface roughness and LOF porosity. Compared to the deterministic simulation with a constant heat source model, the stochastic simulation also provides statistics of the melt pool geometries (depth and width). This useful information can be used for the uncertainty quantification of the process. 

The PPB-DT model, with its stochastic calibration framework, significantly reduces computational effort. In general, multiple-parameter calibration poses a high-dimensional problem that necessitates numerous forward simulations to formulate the optimization problem. Standard approaches such as genetic algorithms [34] have previously been employed for high-dimensional calibration problems, requiring the computational model to repeatedly evaluate a trial set of parameters. This results in a computationally expensive model. Conversely, the PPB-DT model constructs a powerful non-intrusive data-driven HOPGD model for the calibration scheme, accelerating the evolution of the trial set of parameters. HOPGD decomposes an n-dimensional problem into a series of one-dimensional problems. It seeks an $L^2$ projection of data to compute mode functions capable of reproducing or extrapolating the full parametric function. As a result, it serves as a potent surrogate model in the stochastic calibration scheme and significantly reduces the degrees of freedom and computational costs. Having access to more experimental data for calibration can lead to higher prediction accuracy. However, the availability of experimental data is limited, and there is a need for more open data sources, such as NIST AM Bench, to be contributed by the additive manufacturing (AM) community.

In the current model, uncertainty in the additive manufacturing (AM) process is primarily assumed to stem from the parameters of the heat source model, which are modeled with Gaussian distributions. While the current model demonstrates good predictive performance for melt pool dimensions such as width and depth, it may lack the ability to accurately predict other crucial AM indicators, such as solid cooling rate, liquid cooling rate, and time above melting \citep{amin2023physics}. To address this limitation, a more intricate stochastic model, such as a Bayesian model, can be employed to characterize the uncertainty of the heat source model. Bayesian models have the advantage of incorporating prior information about the AM process, resulting in more precise estimations.

The PPB-ML-DT model enables us to make faster predictions for melt pool phenomena. It's important to note that the accuracy of this model depends on several factors, including the choice of machine learning techniques, the quality of the training data, and proper training procedures. In this regard, multi-fidelity data, derived from both experiments and simulations, can be valuable. However, when using experimental data, caution is advised due to the potential presence of noise, which can significantly impact the training process. Using a sophisticated machine learning model may inadvertently focus on fitting the noise rather than accurately representing the underlying physics of the melt pool. In such cases, the PPB-DT model can prove useful, as the data it provides tends to have less noise. However, it may lack certain aspects of the physics if not explicitly considered in the model. To address these challenges and find a balance between different datasets' fidelity while capturing the essential physics for improved prediction, a transfer learning approach can be a valuable tool.

\section{Future directions}
\label{sec6}
While the current method uses the high-fidelity AM-CFD model to simulate the laser powder bed fusion process, the efficiency of AM-CFD is highly constrained by the total number of DoFs (DoFs) in the system. The DoFs increase exponentially as the number of elements in each domain increases. As a result, direct numerical simulation (DNS) of part-scale structures with AM-CFD can be exorbitant. Furthermore, the offline calibration stage may encounter the curse of dimensionality when the dimension of the parametric space is high. To solve these issues, the statistical space-time-parameter tensor decomposition (TD) method can be used as a highly accurate reduced-order method to parameterize spatial, temporal and parametric domains at the same time. Thanks to TD, the DoFs grow linearly with respect to the number of elements in each domain. As a result, we expect to solve large systems efficiently while maintaining high accuracy \citep{li2023convolution}. More importantly, since the parameter space (material parameter, process parameter, boundary condition parameter) is parameterized, the solution at any combination of parameters can be easily obtained by using finite element interpolation without solving the system of equations again. Consequently, the calibration process can be further accelerated.

The current work represents the initial phase of our ongoing effort to develop robust process modeling tools for laser powder bed fusion. In our future endeavors, we plan to expand the stochastic calibration framework to encompass more complex scenarios, including multi-layer and multitrack cases with intricate part geometries and tool paths. We have already manufactured a build plate containing 480 samples with varying part shapes, such as cylinders, squares, L-shapes, squares with holes, tapered L-shapes, and overhangs, all of which were closely monitored during the melt pool formation. This extensive dataset will play a crucial role in our future work, enabling us to further enhance the PPB-DT model and train the PPB-ML-DT model for controlling various melt pool parameters, including temperature, depth, width, aspect ratio, and their influence on resulting microstructure.

\section{Conclusions}
\label{sec7}
In summary, we have developed two digital twin models, the PPB-DT and PPB-ML-DT, for the laser powder bed fusion process. These models have been demonstrated for melt pool prediction, defect diagnosis, online monitoring, and control applications. The PPB-DT model is a stochastically calibrated physics-based model capable of providing melt pool statistics and offering improved accuracy in predicting surface roughness and LOF porosity. In the stochastic calibration process, we have implemented a mechanistic reduced-order HOPGD model with Markov Chain Monte Carlo (MCMC) sampling, which significantly reduces the calibration time. We have validated the PPB-DT model using AFRL multi-layer and multitrack experiments and demonstrated its effectiveness in diagnosing defects in NIST overhang parts. The PPB-DT model provides statistically predicted surface roughness and porosity for part-scale simulations, aligning closely with experimental distributions, all achieved at a considerably reduced computational cost while maintaining high-fidelity computational modeling. The PPB-ML-DT model is employed for controlling melt pool geometries, allowing real-time process control with rapid prediction capabilities. Ultimately, these modeling and simulation tools enable us to make part-scale predictions of defects and offer insights into control strategies for an effective defect mitigation plan and control strategy for the desired performance of the build parts.

\section*{Acknowledgements}
\label{sec:acknowledgement}
 W.K. Liu and G.J. Wagner  would like to acknowledge the support of NSF Grant CMMI-1934367 for up to section \ref{sec3} of the paper. Y.L. Li would like to acknowledge the support of Predictive Science and Engineering Design (PSED) Graduate Program of Northwestern University.

\appendix
\section{Thermal-Fluid analysis of LPBF process with stochastic heat source model}
\label{app1}
To simulate the melted track geometries in the LPBF process, we perform a thermal-fluid analysis using a stochastic heat source model. This model is an extension of our well-tested AM-CFD code and takes into account the thermal analysis of the entire part while the melt pool region is modeled with fluid dynamics and heat transfer. The governing equations for mass, momentum, and energy conservation were derived to solve the thermal-fluid model: \citep{lu2020adaptive}

\begin{equation}
\int_{\Omega f_l}{(\rho\nabla\cdot\bm{u})}dV=0
\end{equation}

\begin{equation}
\int_{\Omega f_l}\left(\frac{\partial(\rho \bm{u})}{\partial(t)}+\nabla\cdot{\rho \bm{u} \bm{u}}-\mu\nabla^2\bm{u}+\nabla p+\frac{180\mu{(1-f_l)}^2}{c^2(f_l^3+B)}\bm{u}-\rho_{0} g\beta(T-T_{0}) \bm{I}\right)dV=0
\label{Eq:A2}
\end{equation}

\begin{equation}
\int_{\Omega}\left(\frac{\partial(\rho h+\rho \Delta H)}{\partial t}+\nabla\cdot(\rho \bm{u} h+\rho \bm{u} \Delta H+\nabla\cdot \bm{q})\right)dV=0
\end{equation}

\noindent where $t$ is the time, $\bm{u}$ is the velocity, $\mu$ denotes the viscosity, $p$ is the pressure,  $T$ is the temperature, $\rho$ is the density, and $\beta$ is the thermal expansion coefficient. $g$ is the acceleration of gravity and equals to 9.8 $m/s^2$.  $\rho_{0}$ and $T_{0}$ are density and temperature of reference material. $H$ is the specific enthalpy, and can be divided into the sum of sensible heat $h$ and the
latent heat of fusion $\Delta H$. In this paper, $\mu$ is set as a constant, $c$ is the approximate primary dendritic spacing, which is set to 1 $\mu$m. $B$ is used to avoid division by zero and set as $10^{-6} m$. $\Omega f_l$ denotes the melt pool region and $\Omega$ is the whole domain. $f_l$ is the volume fraction of the liquid phase, which is defined as:

\begin{equation}
\begin{cases}                                   

                   f_l=0\ \ \ \ \ \ {if} \ \ \  T\leq T_s\\            

                   f_l=\frac{T-T_s}{T_l-T_s}\ \ \ \ \ \ {if} \ \ \  T_s \textless T \textless T_l\\
                   
                   f_l=1\ \ \ \ \ \ {if} \ \ \  T\geq T_l\\

                   \end{cases}
\end{equation}

\noindent where $T_s$ and $T_l$ are the solidus and liquidus temperature of materials, respectively.

Considering $\bar{\bm{q}}$ on the surface boundary, heat flex $\bm{q}$ and its relation with temperature $T$ is 

\begin{equation}
\bm{q}=-\bm{k}\cdot \nabla T
\end{equation}

\noindent where $\bm{k}$ is the thermal conductivity tensor. In isotropic cases, $\bm{k}=k\bm{I}$ denotes the second-order identity tensor. The heat source and boundary condition can be written as:

\begin{equation}
\begin{cases}                                   

                   \bar{\bm{q}}\cdot \bm{n}=h_c(T-T_{0})-\sigma_s\varepsilon(T^4-T_{0}^4)+q_{source} \ \ \ \ \ \    on \ \ \partial\Omega_q\\            

                   T=\bar{T}\ \ \ \ \ \ \ \ \ \ \ \ \ \  \ \ \ \ on \ \ \partial\Omega_q\\             

                   \end{cases}
\end{equation}

\noindent where $h_c$ defines the convective heat transfer coefficient, $\sigma_s$ is the Stefan–Boltzmann constant, $\varepsilon$ is the emissivity, $\bm{n}$ is the normal direction of heat source surface.

The heat source $q_{source}$ from the laser, is described by a cylindrical shape conjugated with Gaussian intensity distribution in AM-CFD program. There are many different heat source models that can be implemented such as cylindrical, semi-spherical, semi-ellipsoidal, conical, radiation heat transfer, ray-tracing, linear decaying, and exponential decaying, which is summarized comprehensively in reference \citep{sharma2023multiphysics}. The reason for the choice of cylindrical heat source is due to its ability to appropriately match that of experimental melt pool depth, width, cooling rates, and time above melting. There are also some physical foundations behind such cylindrical heat sources. The depth of the cylindrical heat source is based on the optical penetration depth (OPD). Depending on the powder particle size and distribution, the OPD varies. Usually, the OPD is defined as the depth where the intensity of the laser energy reduces. As the analysis under investigation is on a bare plate, the penetration of energy is assumed uniform along the OPD and related to the total amount of energy that is going into the system. Based on this evidence, the cylindrical heat source is chosen for the problem. The equation of the cylindrical heat source model has been discussed in section \ref{sec2}

 The boundary condition for Eq.\ref{Eq:A2} at the top surface is equal to the main driving force (i.e. Marangoni force):

 \begin{equation}
\tau_x=\mu\frac{\partial u_x}{\partial z}=\frac{d\gamma}{dT}\nabla_xT
\end{equation}

 \begin{equation}
\tau_y=\mu\frac{\partial u_y}{\partial z}=\frac{d\gamma}{dT}\nabla_yT
\end{equation}

\noindent where $\gamma$ is the surface tension, which depends on both temperature and materials, and $\frac{d\gamma}{dT}$ is the temperature
coefficient.

The powder layer is treated as a continuous media, and it is distinguished from the substrate through its material properties. A consolidated factor $\alpha$ ranging from 0 to 1 is used to identify the material state. The value of 0 stands for the material is in the original powder state (no consolidation), while 1 denotes a bulk state (fully consolidated). The definition of $\alpha$ is:

\begin{equation}
\alpha=\frac{T_{peak}-T_s}{T_l-T_s}
\end{equation}

\noindent where $T_{peak}$ is the local peak energy, and $T_s$\ and $T_l$ are solid and liquid temperature of material, respectively.

The thermo-physical properties of IN625 are summarized in Table \ref{tab:propoerties}. The densities at ambient and liquidus temperatures are used for solid and liquid densities, respectively. Temperature-dependent polynomials were used for the solid’s thermal conductivity and solid’s specific heat capacity.

\begin{table}[htbp]
  \centering
  \caption{Thermo-physical properties of IN625 and process constants \citep{Cox2021AFRLAM,Capriccioli2009FEPF,osti_5337885,valencia2013thermophysical}}
  \resizebox{\textwidth}{50mm}{
    \begin{tabular}{|p{13em}|c|p{12em}|c|}
    \hline
    Property/parameter & Value & Property/parameter & Value \\

    \hline
    Solid density ($kg\cdot m^{-3}$) & 8440  & Convection coefficient ($W \cdot m^{-1}\cdot K^{-1}$) & 10 \\

    \hline
    Liquid density ($kg\cdot m^{-3}$) & 7640  & Latent heat of fusion ($KJ\cdot kg^{-1}\cdot K^{-1}$) & \multicolumn{1}{c|}{290} \\

    \hline
    Powder density ($kg\cdot m^{-3}$) & 4330  & Dynamic viscosity ($Pa\cdot s$) & $7 × 10^{-3}$ \\

    \hline
    Solidus temperature ($K$) & 1563  & Thermal expansivity ($1/K$) & $5 × 10^{-5}$ \\

    \hline
    Liquidus temperature ($K$) & 1623  & Surface tension ($N\cdot m^{-1}$) & \multicolumn{1}{c|}{1.8} \\

    \hline
    Solid specific heat capacity ($J\cdot kg^{-1}\cdot K^{-1}$) & \multicolumn{1}{p{8em}|}{0.2441$T$ + 338.39} & Marangoni coefficient ($N\cdot m^{-1}\cdot K^{-1}$) & -$3.8 × 10^{-4}$ \\

    \hline
    Liquid specific heat capacity ($J\cdot kg^{-1}\cdot K^{-1}$) & 709.25 & Emissivity & \multicolumn{1}{c|}{0.4} \\

    \hline
    Powder specific heat capacity ($J\cdot kg^{-1}\cdot K^{-1}$) & \multicolumn{1}{p{8em}|}{0.2508$T$ + 357.70} & Ambient temperature ($K$) & \multicolumn{1}{c|}{295} \\

    \hline
    Solid thermal conductivity ($W\cdot m^{-1}\cdot K^{-1}$) & \multicolumn{1}{p{8em}|}{0.0163$T$ + 4.5847} & Reference temperature ($K$) & \multicolumn{1}{c|}{295} \\

    \hline
    Liquid thermal conductivity ($W\cdot m^{-1}\cdot K^{-1}$) & 30.078 & Preheat temperature ($K$) & \multicolumn{1}{c|}{353} \\

    \hline
    Powder thermal conductivity ($W\cdot m^{-1}\cdot K^{-1}$) & 0.995 & Stefan–Boltzmann constant ($W\cdot mm^{-2}\cdot K^{-4}$) & $5.67 × 10^{-14}$ \\
    \hline

    \end{tabular}%
  \label{tab:propoerties}}%
\end{table}%

In order to consider the influence of the localized preheating from adjacent scan paths that leads to transient behavior of the vapor depression, the residual heat factor (RHF) is considered into the heat source model \citep{yeung2020residual}. RHF at specific point $i$ is defined as:

 \begin{equation}
{RHF}_i=\sum_{k\in S_i}\left(\frac{R-d_{ik}}{R}\right)^2 \left(\frac{T-t_{ik}}{T}\right) L_k
\end{equation}

\begin{figure}[h!]
\centering
\includegraphics[width=0.4\textwidth]{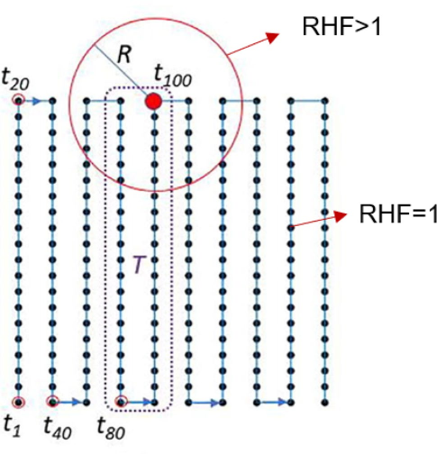}
\caption{Quantitative comparisons of cross-sectional area between experimental measurements and stochastic simulation \citep{yeung2020residual}}
\label{fig:scanpath}
\end{figure}

The scan path consists of discrete points determined by the simulation's time step and the laser scan speed. Distance between the point $i$ and $k$, denoted as $d_{ik}$, represents the preheating effect of point $k$ on point $i$. Similarly, the elapsed time since point $k$ was scanned is denoted as $t_{ik}$. The normalized laser power at point $k$, denoted as $L_k$, is 1 when the laser is on and 0 when laser is off. Constants $R$ and $T$ have values of $2\times {10}^{-4}$ and $2\times {10}^{-3}$, respectively. These constants act as thresholds to exclude points that have not interacted with the laser for a sufficient amount of time. Points within the threshold belong to set $S_i$, defined as $S_i=\{t_{ik}<T\cup\ d_{ik}<R,\ \ where\ i>k\}$. The RHF is normalized as $RHF=\frac{{RHF}_i}{{RHF}_c}$, where ${RHF}_c$ equals to ${RHF}_i$ at the middle part of the toolpath and $RHF$ is greater than 1 at the corner of the toolpath, as shown in Figure \ref{fig:scanpath}. 

Considering the influence of residual heat, the heat source parameters can be coupled with the RHF as:

 \begin{equation}
 d=P1\frac{P}{V}{\rm RHF}^2
\end{equation}

 \begin{equation}
\eta=\max(P2\frac{P}{V}{\rm RHF}^2,0.28)
\end{equation}

 \begin{equation}
 r_b=P3\frac{P}{V}{\rm RHF}^2
\end{equation}

Again, the three unidentified parameters $P1,P2,P3$ require calibration. The calibration process for these parameters will be discussed in the following section.

\section{Kernel Density Estimation}
\label{app2}
Kernel Density Estimation (KDE) \citep{Davis2011} is a powerful method for estimating the probability density function of a random variable. KDE is used to find the distribution of experimental measurements which is further utilized to calibrate the stochastic simulation process parameters. The density function can be estimated by taking the first derivative of the distribution function. One of the simplest and most effective techniques for estimating distribution functions is the Empirical Distribution Function (EDF). In this method, the estimate of $F_n\left(t\right)$ is the probability of all samples less than $t$:

 \begin{equation}
F_n\left(t\right)=\frac{The\ number\ of\ elements\ in\ the\ sample\ \le t}{n}=\frac{1}{n}\sum_{i=1}^{n}\mathbf{1}_{wi\le t} \label{eq22}
\end{equation}

\noindent where $\mathbf{1}_{wi\le t}$ is the indicator function. The indicator function of a subset A of a set $W$ is defined as:

\begin{equation}
\mathbf{1}_A\left(w\right)=\begin{cases}                                   

                   1\ \ \ if\ w\in\ A\\            

                   0\ \ \ if\ w\notin\ A\\             

                   \end{cases} \label{eq23}
\end{equation}

EDF is not differentiable and not smooth enough to compute the density function by the first derivative with respect to EDF. Thus, central difference can be used to find the density function:

\begin{equation}
f(w)=\lim \limits_{h\rightarrow 0}{\frac{F\left(w+h\right)-F(w-h)}{2h}} \label{eq24}
\end{equation}

Replace the distribution function with the empirical distribution function in Eq.\ref{eq22}, the numerator of Eq.\ref{eq24} is the number of points falling in the interval $[w-h,\ w+h]$, which can be written as:

\begin{equation}
f\left(w\right)=\frac{1}{2Nh}\sum_{i=1}^{N}{\mathbf{1}(w-h\le w_i\le w-h)}=\frac{1}{Nh}\sum_{i=1}^{N}{\frac{1}{2}\ast\mathbf{1}(\frac{\left|x-x_i\right|}{h}\le1)} \label{eq25}
\end{equation}

\noindent where $N$ is the number of sample points and h is the bandwidth. If so-called kernel function $K\left(t\right)=\frac{1}{2}\ast{\mathbf{1}}(t\le1)$ is used, Eq. \ref{eq25} can be further written as:

\begin{equation}
f\left(w\right)=\frac{1}{Nh}\sum_{i=1}^{N}{K(\frac{w-w_i}{h})} \label{eq26}
\end{equation}

Eq.\ref{eq26} gives the expression of KDE, which is also the estimation of probability density function. The integration of Eq.\ref{eq26} is:

\begin{equation}
\int{f\left(w\right)dw=\frac{1}{Nh}\sum_{i=1}^{N}\int{K(\frac{w-w_i}{h})}dw}=\frac{1}{N}\sum_{i=1}^{N}\int{K\left(t\right)dt=}\int K\left(t\right)dt
\end{equation}

Thus, as long as the integration of K is equal to 1, the integration of the estimated density function can be guaranteed as 1. The standard normal distribution can be used as kernel function, whose expression is:

\begin{equation}
K\left(t\right)=\frac{1}{\sqrt2\pi}e^{-\frac{t^2}{2}}
\end{equation}

Notice that the choice of h (also called bandwidth) in Eq.\ref{eq26} influence the goodness of KDE model. Here, Silverman’s rule of thumb algorithm is used as bandwidth selector due to its universality and effectiveness:

\begin{equation}
h=0.9\ast\ min\ {(\hat{\sigma},IQR/1.35)N}^{-\frac{1}{5}}
\end{equation}

\noindent where $\hat{\sigma}$ is the standard deviation of samples and $IQR$ is interquartile range (the difference between 75th and 25th percentiles).

\section{Kullback-Leibler Divergence for statistical test}
\label{app3}
Since both experimental and simulation results are distributions $(f_{We}({{x}}), f_{De}\left({x}\right),\\f_{Ws}({x}), f_{Ds}({x}))$, Kullback-Leibler Divergence (KLD) can be used here to express difference between two continuous probability density distributions. KLD has its origins in the entropy of information theory \citep{10.1214/aoms/1177729694}, typically denoted as $H$. The definition of entropy for a probability distribution is:

\begin{equation}
H=-\int_{x}{p(x)\log(p\left(x\right))}dx
\end{equation}

\noindent where $p(x)$ is the probability density function of any random $x$. With the help of entropy, information can be quantified, and the loss of information can be measured when the observed distribution is substituted with parameterized approximation. Similarly, rather than just having probability distribution $p(x)$, KLD adds in the approximating distribution $q\left(x\right)$ and takes logarithmic operation:

\begin{equation}
D_{KL}(p\left(x\right),q\left(x\right))=-\int_{x}{p(x)\log(p\left(x\right)-q\left(x\right)) }dx
\end{equation}

Essentially, KLD is the expectation of the log difference between the original (experimental) distribution with the approximating (simulated) distribution. A more common way to see KL divergence written is as follows:

\begin{equation}
D_{KL}(p\left(x\right),q\left(x\right))=\int_{x}{p(x)\log(\frac{q(x)}{p(x)})}dx
\end{equation}

\section{Markov chain Monte Carlo (MCMC) algorithm}
\label{app4}

Markov chain Monte Carlo (MCMC) algorithms, like the renowned Metropolis-Hastings algorithm (\citep{hastings1970monte}), and the Gibbs sampler (e.g., Geman and Geman \citep{geman1993stochastic}; Gelfand and Smith \citep{gelfand1990sampling}), have gained immense popularity in the field of statistics. They provide a powerful means to obtain approximate samples from complex probability distributions in high-dimensional spaces. The impact of MCMC algorithms on Bayesian inference has been truly remarkable, enabling practitioners to effectively sample from posterior distributions of intricate statistical models. In this project, the MCMC is used to generate samples for the stochastic AM simulation based on the calibrated process parameters.

The core problem tackled by MCMC algorithms can be described as follows. We are presented with a density function $\pi_u$ defined on a particular state space $\mathcal{X}$. Although $\pi_u$ may not be normalized, it satisfies the condition $0<\int_{\mathcal{X}} \pi_u<\infty$. Typically, $\mathcal{X}$ represents an open subset of $\mathbf{R}^d$, and the densities are evaluated using Lebesgue measure. However, other scenarios, such as discrete state spaces, are also feasible. This density function gives rise to a probability measure $\pi(\cdot)$ on $\mathcal{X}$, accomplished through a process of transformation.

\begin{equation}\pi(A)=\frac{\int_A \pi_u(x) d x}{\int_{\mathcal{X}} \pi_u(x) d x}
.\end{equation}

Our objective is to estimate expectations of functions $f: \mathcal{X} \rightarrow \mathbf{R}$ with respect to the probability measure $\pi(\cdot)$. In other words, we aim to estimate the value of $\pi(f)$, which can be defined as the expected value of $f(X)$ under $\pi(\cdot)$. Mathematically, it is represented as:
\begin{equation}\pi(f)=\mathbf{E}\pi[f(X)]=\frac{\int{\mathcal{X}} f(x) \pi_u(x) d x}{\int_{\mathcal{X}} \pi_u(x) d x} .
\end{equation}
When the state space $\mathcal{X}$ is of high dimensionality, and the density function $\pi_u$ is complex, direct integration of the integrals in Equation (D.2) becomes impractical, both analytically and numerically.

The conventional Monte Carlo approach to address this problem involves simulating independent and identically distributed (i.i.d.) random variables $Z_1, Z_2, \ldots, Z_N \sim \pi(\cdot)$. Subsequently, we can estimate $\pi(f)$ by utilizing the following formula:

\begin{equation}\hat{\pi}(f)=(1 / N) \sum_{i=1}^N f\left(Z_i\right).
\end{equation}

This approach provides an unbiased estimate with a standard deviation of approximately $O(1 / \sqrt{N})$. Additionally, if $\pi\left(f^2\right)<\infty$, the classical Central Limit Theorem ensures that the error $\hat{\pi}(f)-\pi(f)$ will converge to a normal distribution, which is beneficial for analysis. However, a significant challenge arises when $\pi_u$ is a complex function, making it exceedingly difficult to directly generate i.i.d. random variables from $\pi(\cdot)$.

The solution provided by Markov chain Monte Carlo (MCMC) is to construct a Markov chain on the state space $\mathcal{X}$ that can be easily simulated on a computer and has $\pi(\cdot)$ as its stationary distribution. In other words, our goal is to define transition probabilities $P(x, d y)$ for $x, y \in \mathcal{X}$ in a way that the following equation holds:

\begin{equation}\int_{x \in \mathcal{X}} \pi(d x) P(x, d y)=\pi(d y) .
\end{equation}

By running the Markov chain for a sufficiently long time (regardless of the starting point), the distribution of $X_n$ will become approximately stationary: $\mathcal{L}\left(X_n\right) \approx \pi(\cdot)$. Consequently, we can set $Z_1=X_n$, restart and rerun the Markov chain to obtain $Z_2, Z_3$, and so on. These obtained values can then be used to perform estimates.


 \bibliographystyle{elsarticle-num} 
 \bibliography{main}





\end{document}